\documentclass[runningheads]{llncs}

\usepackage{eccv}
\usepackage{eccvabbrv}

\usepackage{graphicx}
\usepackage{booktabs}

\usepackage[accsupp]{axessibility}  
\usepackage[mode=buildnew]{standalone}

\usepackage{hyperref}

\usepackage{orcidlink}

\usepackage{amsmath,amsfonts,bm}

\def\eqref#1{equation~\ref{#1}}

\def\1{\bm{1}}

\DeclareMathAlphabet{\mathsfit}{\encodingdefault}{\sfdefault}{m}{sl}
\SetMathAlphabet{\mathsfit}{bold}{\encodingdefault}{\sfdefault}{bx}{n}

\newcommand{\R}{\mathbb{R}}

\DeclareMathOperator*{\argmax}{arg\,max}

\usepackage{amsmath}
\usepackage{bm}
\usepackage{amssymb}
\usepackage{booktabs}
\usepackage{comment}
\usepackage{multirow}
\usepackage{multicol}
\usepackage{makecell}
\usepackage{epsfig}
\usepackage{color}
\usepackage{soul}
\usepackage{xspace}
\usepackage{tikz}
\usepackage{appendix}
\usepackage{rotating}
\usepackage{pdflscape}
\usepackage{cite}
\usepackage{url}
\usepackage{caption}
\usepackage{indentfirst}
\usepackage{fontawesome5}
\newcommand{\OURS}{NOVUM\xspace}

\definecolor{darkgreen}{rgb}{1, 0.15, 0.15}
\definecolor{revThree}{rgb}{1, 0.5, 0}
\definecolor{revTwo}{rgb}{0.73, 0.29, 0.77}

\newcommand{\beginsupplement}{%
    \setcounter{table}{0}
    \renewcommand{\thetable}{S\arabic{table}}%
    \setcounter{figure}{0}
    \renewcommand{\thefigure}{S\arabic{figure}}%
 }
\usepackage{amsmath}
\usepackage{subcaption}
\usepackage{enumerate}
\usepackage{enumitem}
\usepackage{subcaption}
\newcommand\ddfrac[2]{\frac{\displaystyle #1}{\displaystyle #2}}

\begin{document}

\title{\OURS: Neural Object Volumes for \\ Robust Object Classification}

\makeatletter
\def\thanks#1{\protected@xdef\@thanks{\@thanks
        \protect\footnotetext{#1}}}
\makeatother

\newcommand*\samethanks[1][\value{footnote}]{\footnotemark[#1]}

\author{
Artur Jesslen\inst{1}\thanks{$^\star$ \ Equal contribution.}\samethanks[1]\textsuperscript{\faEnvelope[regular]}\thanks{\textsuperscript{\faEnvelope[regular]} Corresponding author: \email{jesslen@cs.uni-freiburg.de}.}\orcidlink{0000-0002-4837-8163}
\and
Guofeng Zhang\inst{2}\samethanks[1]\orcidlink{0009-0005-8183-0771} 
\and
Angtian Wang\inst{2}\orcidlink{0009-0006-9189-5277}
\and
Wufei Ma\inst{2}\orcidlink{0000-0002-4696-2833}
\and\\
Alan Yuille\inst{2}\orcidlink{0000-0001-5207-9249} 
\and
Adam Kortylewski\inst{1,3}\orcidlink{0000-0002-9146-4403}
}

\authorrunning{A.~Jesslen et al.}

\institute{
University of Freiburg
\and
Johns Hopkins University
\and
Max-Planck-Institute for Informatics}

\def\ie{\emph{i.e.,~}}
\def\eg{\emph{e.g.,~}}

\maketitle
\begin{abstract}
Discriminative models for object classification typically learn image-based representations that do not capture the compositional and 3D nature of objects. 
In this work, we show that explicitly integrating 3D compositional object representations into deep networks for image classification leads to a largely enhanced generalization in out-of-distribution scenarios.
In particular, we introduce a novel architecture, referred to as \OURS,  that consists of a feature extractor and a \textit{neural object volume} for every target object class. 
Each neural object volume is a composition of 3D Gaussians that emit feature vectors.
This compositional object representation allows for a highly robust and fast estimation of the object class by independently matching the features of the 3D Gaussians of each category to features extracted from an input image.
Additionally, the object pose can be estimated via inverse rendering of the corresponding neural object volume. 
To enable the classification of objects, the neural features at each 3D Gaussian are trained discriminatively to be distinct from (i) the features of 3D Gaussians in other categories, (ii) features of other 3D Gaussians of the same object, and (iii) the background features. 
Our experiments show that \OURS offers intriguing advantages over standard architectures due to the 3D compositional structure of the object representation, namely: (1) An exceptional robustness across a spectrum of real-world and synthetic out-of-distribution shifts and (2) an enhanced human interpretability compared to standard models, all while maintaining real-time inference and a competitive accuracy on in-distribution data.
Code and model can be found at 
\href{https://github.com/GenIntel/NOVUM}{\faGithub/GenIntel/NOVUM}.


\end{abstract}

\section{Introduction}
\begin{figure}
\centering
    \includegraphics[width=0.9\linewidth]{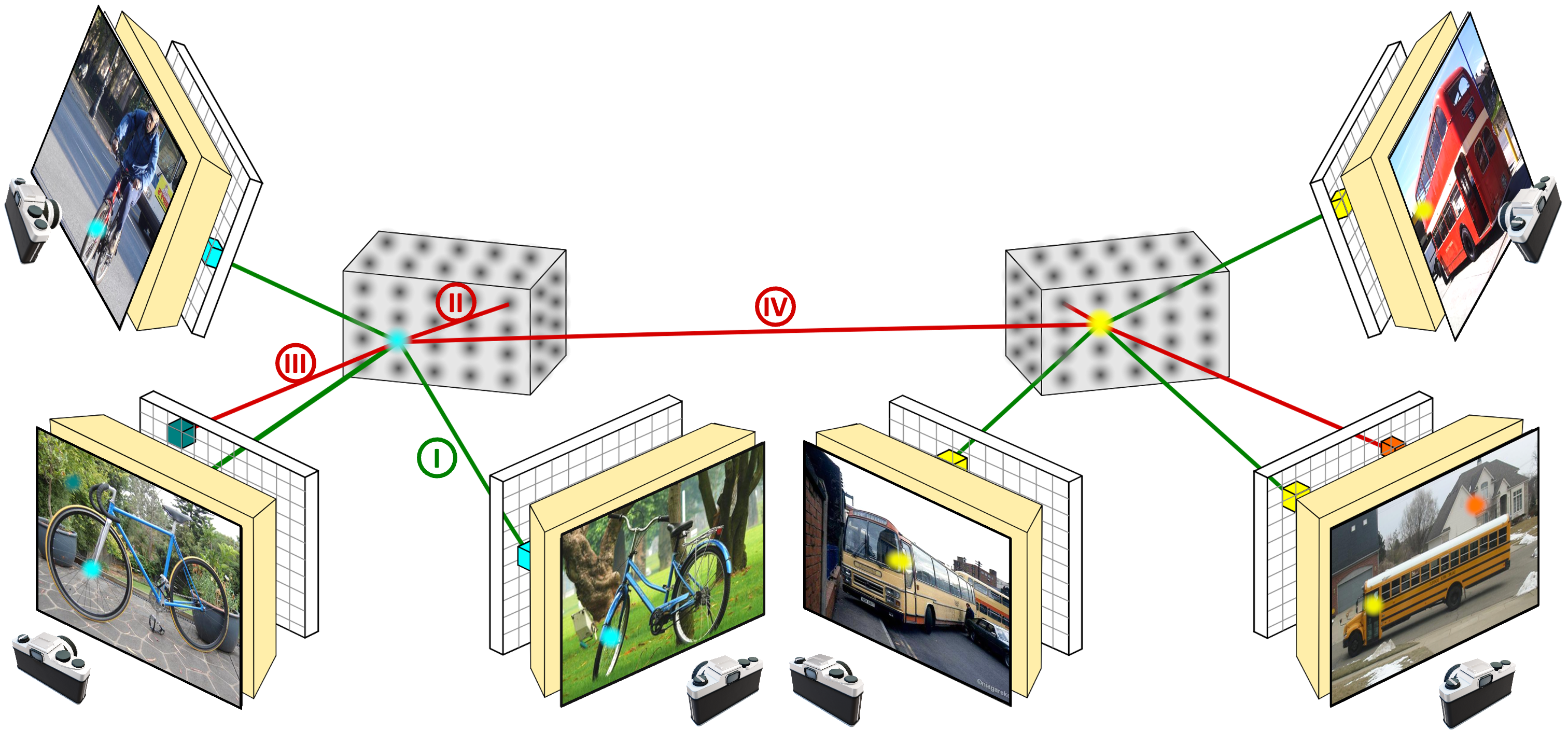}
    \caption[]
    {Schematic overview of how \OURS is trained. 
    The model consists of a shared backbone (yellow) and one neural object volume for each object class (grey), which are represented as 3D Gaussians on a cuboid shape.
    During training, the backbone first computes feature maps of the training images.
    Given the class label and the 3D object pose, the backbone is trained in a contrastive manner using four types of losses: (I) To make features of the same Gaussian similar across instances (green), while at the same time making the features distinct (red) from (II) features of Gaussians from the same object, (III) background features, and (IV) features of Gaussians from other objects.
    } 
    \label{fig:teaser}
\end{figure}
Current deep learning architectures demonstrate advanced capabilities in visual recognition tasks, \eg object classification, detection, and pose estimation \cite{lecun1995convolutional,simonyan2014very,he2016deep,dosovitskiy2020image,liu2021swin}. 
However, generalization to out-of-distribution (OOD) scenarios remains a fundamental challenge \cite{hendrycks2019benchmarking,michaelis2019corruptions,zhao2022ood, kortylewski2020compositional}. 
In contrast, human vision achieves a significantly better robustness under OOD scenarios, e.g., domain shift, and occlusions \cite{bengio2021deep,kortylewski2020combining}.
Some cognitive studies hypothesize that human vision relies on a compositional 3D representation of objects while perceiving the world through an analysis-by-synthesis process \cite{neisser1967cognitive,yuille2006vision}. 
While object-centric  \cite{yi2018neural,yi2019clevrer,locatello2020object} and compositional representations \cite{jin2006context,george2017generative,kortylewski2020compositional} show promise in improving sample efficiency and generalization of machine learning algorithms, so far these models are largely ignorant of the 3D nature of our world and mostly limited to simple synthetic domains.
This raises the question: Can machines enhance generalization at real-world classification tasks through learning 3D object representations?
In this work, we embed a 3D compositional object representation explicitly into the neural network architecture for object classification. We take inspiration from recent advances in neural rendering \cite{kerbl20233d} and prior works on pose estimation \cite{iwase2021repose,wang2021nemo} to design a 3D-aware neural network for image classification.
In particular, we propose \OURS, which is composed of a feature extractor and \textit{neural object volumes} for every object category (\Cref{fig:teaser}). 
Each neural object volume is a spatial composition of Gaussian ellipsoidal kernels that emit feature vectors.
During inference, an image is classified by matching the Gaussian features of each category to the input feature map. 
In this way, object classification is realized using the individual Gaussian features only and without requiring the 3D spatial geometry, hence facilitating a fast and robust classification inference. 
Each Gaussian contributes to the prediction by focusing individually on different evidences (different parts of objects in our case). When combined together, the lack of evidence from part of the Gaussians can be compensated by the rest, leading to high robustness against outliers or occlusions.
Moreover, the inherent 3D representation in our model can also estimate the 3D object pose via inverse rendering of the feature volume using Gaussian splatting.
We use 3D pose annotations of the objects to train the neural features to be distinct from features of other categories' neural object volumes, as well as spatially apart features of the same neural object volume, and the background clutter. 
Intuitively, the feature extractor learns to classify every pixel in the image as being either part of the object or as background context.

We evaluate \OURS on a variety of datasets that contain 3D pose annotations and OOD shifts, such as real-world OOD shifts on the OOD-CV dataset \cite{zhao2023ood}, and synthetic OOD shifts on the corrupted PASCAL3D+ \cite{hendrycks2019benchmarking} and occluded PASCAL3D+\cite{wang2021nemo}.
Our experiments show that \OURS is exceptionally robust compared to other state-of-the-art architectures (both CNNs and Transformers) at object classification while performing on par with in-distribution data in terms of accuracy and inference speed. 
Moreover, the 3D pose predictions obtained via inverse rendering are competitive to baseline models which are limited to perform 3D pose estimation only.
Finally, we show that the Gaussian matching result provides intuitive human interpretable information about the model prediction, by showing where the model perceives corresponding object parts. 

In conclusion, \OURS introduces a 3D volume representation for classification, achieving robustness through compositionality and 3D-awareness, while offering enhanced human interpretability by visualizing individual Gaussian kernel matches, and still enabling real-time inference.

\section{Related Work} \label{sec:related}

\noindent \textbf{Robust Image Classification.} 
Image classification is a classical task in computer vision. 
Multiple influential architectures such as ResNet \cite{he2016deep}, some transformers\cite{vaswani2017attention,liu2021swin} were designed for this task. 
However, these models have primarily targeted in-distribution data, leaving a significant gap when faced with out-of-distribution (OOD) data such as common synthetic corruptions \cite{hendrycks2019benchmarking} or real-world OOD data \cite{zhao2022ood}. 
To bridge this OOD generalization gap, efforts have concentrated mainly on two fronts: data augmentation and architectural design. 
Data augmentation strategies involve leveraging learned augmentation policies \cite{cubuk2018autoaugment}, and data mixtures \cite{hendrycks2019augmix} to enhance the diversity of training samples, thereby fostering generalization by generating synthetic OOD data. 
On the other hand, architectural advances that incorporate general prior knowledge about the world, such as compositionality \cite{jin2006context,george2017generative,kortylewski2020compositional}, or object-object centric representations \cite{yi2018neural,yi2019clevrer,locatello2020object}, have also shown promising advances in terms of enhancing sample efficiency and generalization in neural networks. 
Our approach falls into the second category, as we introduce a 3D compositional representation into the architecture of neural networks for object classification. 
In contrast to standard discriminative methods relying on a single entangled feature representation of an image, our 3D and compositional representation leads to a largely enhanced robustness when faced with occlusions, corruptions, and real-world OOD scenarios.\\
\noindent \textbf{Contrastive learning.} 
Studies have found that, in supervised settings,  learning features that are discriminative during training does not guarantee that a model will generalize, instead, they can have inductive biases towards learning simple “shortcut” features and decision rules \cite{hermann2020shapes, nguyen2021wide}. Hence, contrastive learning \cite{contrastive_base} is an interesting direction to prevent to learn these shortcuts since it is essentially a framework that learns similar/dissimilar representations by optimizing through the similarities of pairs in the representation space. Later the idea of using contrastive pair is extended to Triplet \cite{triplet}. While traditional contrastive learning has focused on image-level or text-level representations, the extension to feature-level representations has gained lots of attention after InfoNCE \cite{oord2019representation}. Influential works in various fields include SimCLR \cite{chen2020simple}, CLIP \cite{radford2021learning}, CoKe \cite{bai2022coke}. In our paper, we adopt a three-level feature contrastive loss that encourages spatially distinct features, categorical specific features, and background specific features.\\
\noindent \textbf{3D neural representations for pose estimation.} Explicit 3D object representations can offer significant advantages in terms of generalization over purely image-based representations.
Importantly, when using category-level 3D representations for vision tasks, it becomes imperative to not rely on detailed instance-specific features (\eg object colour or texture), but rather to enable the learning of general category-level features.
As a result, employing features that remain invariant to such specificities emerges as a straightforward solution.
Initially, \cite{stark2010} delved into computing correspondences between image features and a learned volumetric representation, leveraging HOG features and employing a Proposal-Validation process,  which is slow. Later works including \cite{wang2021nemo, iwase2021repose, wang2021neural}, have replaced HOG features with neural features capable of encoding richer information. This shift to neural features has facilitated the extension of render-and-compare methods, originally introduced at the pixel-level \cite{wang2019normalized, chen2020category}, to the feature level. This extension enhances the models' ability to generalize to rendering category-level instances. Conceptually, this method embodies an approximate analysis-by-synthesis approach, similar to the principles outlined in \cite{grenander1970unified}, which proves to be more robust against out-of-distribution (OOD) data in 3D pose estimation when contrasted with classical discriminative methods \cite{tulsiani2015viewpoints, mousavian20173d,zhou2018starmap}.

\OURS builds on prior work on category-level pose estimation, such as neural mesh models \cite{wang2021nemo,wang2021neural}. \OURS extends this line of work in multiple ways: (1) An architecture with a shared class-agnostic backbone, which enables us to introduce a class-contrastive loss for object classification.
(2) A principled, efficient way of performing classification inference that exploits individual 3D Gaussians.
(3) A comprehensive mathematical formulation that derives a vMF-based contrastive training loss. 
(4) A compositional 3D Gaussian representation~\cite{kerbl20233d} that can be matched robustly with volume rendering \cite{wang2022voge} even in strong OOD scenarios. 
We demonstrate the applicability of compositional 3D representations in classification tasks, hence pioneering a step towards highly advanced
capabilities in terms of generalization, interpretability and multi-tasking in vision.

\section{Method}
\label{sec:method}
In this section, we present a deep network with an integrated 3D and compositional volume representation of objects to achieve robust object classification (\Cref{sec:method:representation}).
We discuss how the model can be learned (\Cref{sec:method:training}) and how it can be applied to infer the object class as well as the 3D pose at test time (\Cref{sec:method:inference}).

\subsection{\OURS: A Network Architecture with Neural Object Volumes}
\label{sec:method:representation}

\noindent \textbf{Motivation.} Our aim is to achieve robustness at object classification by devising a neural network architecture that integrates a 3D compositional representation of objects.
Specifically, our architecture explicitly uses an object-centric 3D representation to introduce compositionality on two levels of abstraction: (1) \textit{Image-level compositionality}, \ie a representation that explicitly models an image as a composition of objects and background clutter. (2) \textit{Object-level compositionality}, \ie a representation that explicitly models objects as a spatial composition of local elements. 
This compositional representation enhances model robustness by improving the models ability to classify objects even if only few local parts are recognizable (\eg due to occlusion or due to a novel object topology), or when the object is placed in an unusual context (\eg a bicycle underwater).

\noindent \textbf{Neural Object Volumes.}
Building on recent advances in Gaussian splatting~\cite{kerbl20233d}, we represent objects as a 3D density field via a spatial composition of $K$ Gaussians that are placed on the surface geometry of each object category.
Each Gaussian emits an associated feature vector, hence defining the volumetric object representation that we refer to as \textit{neural object volume}. Each neural object volume represents an object category and is learned from feature maps extracted from 2D images by given an annotated 3D pose of the object (\Cref{fig:teaser}). 

More formally, we define a neural volume density at spatial location $\bm{x}\in \mathbb{R}^{3}$ as a mixture of three-dimensional Gaussian $\rho (\bm{x}) = \sum_{k=1}^K \bm{\rho}_k(\bm{x})$. 
Each Gaussian density is defined as $\rho_k(\bm{x}) = \mathcal{N}(\bm{\mu}_k, \bm{\Sigma}_{k})$, where
$\bm{\mu_k} \in \mathbb{R}^{3}$ is the 3D position of the $k$-th Gaussian and $\bm{\Sigma}_k \in \mathbb{R}^{3\times 3}$ is its covariance matrix (defining the direction, shape and size of $k$-th kernel). In our setup, we do not assume detailed geometry of the object but simply arrange the Gaussians such that they form a cuboid volume with a pre-defined and diagonal covariance, such that the volume approximately encompasses the variable object instances in the corresponding object category.
Each Gaussian is associated with a feature vector, denoted $C_k \in \mathbb{R}^{D}$. For each object category $y$, we learn a set of features $\mathcal{C}_y = \{C_k \in \mathbb{R}^D\}_{k=1}^{K}$. 
We define the set of Gaussian features from all object categories as $\mathcal{C} = \{  \mathcal{C}_y \}_{y=1}^Y$, where $Y$ is the total number of object categories. 
The neural object volume can be rendered into the feature space, using standard volume rendering:
{\small  
\begin{align}\label{eq:volume}
    \bm{\hat{C}}_i(\alpha) = \int_{t_n}^{t_f} T(t) \sum_{k=1}^K \rho_k(\bm{r}_{\alpha}(t))\bm{C_k} \mathrm{d}t,\quad\text{where}\ T(t)=\exp\left(-\int_{t_n}^{t} \rho(\bm{r}_{\alpha}(s))\mathrm{d}s\right),
\end{align}
}
where the feature $\bm{\hat{C}}_i(\alpha)$ at pixel position $i$ in the rendered feature map is computed by aggregating the Gaussian features along the ray $\bm{r}_{\alpha}(t)$. 
The ray traverses from the camera center through the pixel $i$ on the image plane with $\alpha$ denoting the camera view.
Here, $t$ ranges from the near plane $t_n$ to the far plane $t_f$. 
The remainder of the image that is not covered by the rendered object volume is represented as background features $\mathcal{B}=\{\beta_n \in \R^D\}_{n=1}^{N_b}$ where $N_b$ is a fixed hyperparameter, and $\mathcal{B}$ is shared among all object categories. 

\noindent \textbf{\OURS.} Our model architecture builds on a feature extractor $\Phi_w$ and a set of neural object volumes, one for each object category. 
The feature extractor computes a feature map $F = \Phi_w(I)  \in \mathbb{R}^{D \times H\times W}$ from an input image $I$, where $w$ denotes the parameters of the CNN backbone. 
The feature map $F$ contains feature vectors $f_i \in \mathbb{R}^{D}$ at positions $i$ on a 2D lattice.

Learning our model requires obtaining correspondences between a Gaussian $k$ of a neural object volume and a location $i$ in the feature map of a training image, and we obtain this correspondence by projecting Gaussian into the image feature map $F$. With given camera pose $\alpha$, we use volume rendering to compute the contribution $\gamma_{ik}$ of Gaussian feature $C_k$ to image features $f_i$. We estimate a one-to-one correspondence between features and Gaussians by selecting the closest image feature $f_i$ for each Gaussian, \ie where $\gamma_{ik}$ is the largest. Throughout the remaining paper, we denote $f_{k \rightarrow i}$ to indicate the extracted feature $f_i$ at location $i$ that Gaussian $k$ with mean $\mu_k$ projects to.
We relate the extracted image features to the Gaussians and background features by von-Mises-Fisher (vMF) probability distributions. 
In particular, we model the probability of generating the feature $f_{i}$ from corresponding Gaussian ${C}_k$ as $P(f_{k \rightarrow i} | {C}_k) = c_M(\kappa) e^{\kappa f_{k \rightarrow i} \cdot {C}_k}$, where ${C}_k$ represents the mean of each vMF distribution ($\lVert f_{k \rightarrow i} \lVert = 1, \lVert {C}_k \lVert = 1$). 
We also model the probability of generating the feature $f_i$ from background features as $P(f_i | \beta_n) = c_M(\kappa) e^{\kappa f_i \cdot \beta_n} $ for $\beta_n \in \mathcal{B}$. 
The concentration parameter $\kappa$, which determines the spread of the distribution and can be interpreted as an inverse temperature parameter, is defined as a global hyperparameter. Hence, the normalization constant $c_M(\kappa)$ is a constant and can be ignored during learning and inference.

\subsection{Learning Discriminative 3D Volume Representations}
\label{sec:method:training}

Learning \OURS is challenging because we not only need to maximize the likelihood functions $P(f_{k \rightarrow i} | {C}_k)$ and $P(f_i|\mathcal{B})$, but also learn a corresponding parameters $w$ of the backbone. 
In particular, we maximize the probability that any extracted feature $f_{k \rightarrow i}$ was generated from $P(f_{k \rightarrow i}| {C}_k)$ instead of from any other alternatives. 
This motivates us to use contrastive learning where we compare the probability that an extracted feature $f_{k \rightarrow i}$ is generated by the correct Gaussian ${C}_k$ or from one of three alternative processes, namely, (i) from the Gaussians of other object classes, (ii) from non-neighboring Gaussians of the same object, and (iii) from the background features (see illustration in \Cref{fig:teaser}):
\begin{align}
\frac{P(f_{k \rightarrow i} | {C}_k) }
    {\sum_{\substack{C_l \in \mathcal{C}_y \\ C_l\notin \mathcal{N}_k}} P(f_{k \rightarrow i} | C_l)  + \omega_{\beta} \sum_{\beta_n \in \mathcal{B}}   P(f_{k \rightarrow i} | \beta_n)  + \omega_{\Bar{y}} \sum_{C_m \in \mathcal{C}_{\Bar{y}}} P(f_{k \rightarrow i} | C_m)},
\end{align}
where $\mathcal{N}_k = \{C_r : \lVert \bm{\mu}_k - \bm{\mu}_r \rVert < \delta, k \neq r \}$ is the neighborhood of $C_k$ and $\delta$ is a distance threshold controlling the size of neighborhood. $y$ is the category of the image and $\Bar{y}$ is a set of all other categories except $y$. $\omega_\beta = \frac{P(\beta_n)}{P(C_k)}$ is the ratio of the probability that an image feature corresponds to the background instead of the Gaussian $k$, and $\omega_{\Bar{y}}  = \frac{P(C_m)}{P(C_k)}$ is the ratio of the probability that an image feature corresponds to Gaussians of other categories instead of the Gaussian $k$. 
We compute the final loss $\mathcal{L}( \mathcal{C}, \mathcal{B}, w)$ by taking the logarithm and summing over all training examples -- all sets of features $\{f_{k \rightarrow i} \}$ from the training set 
\begin{equation}
     -\sum_y\sum_{k=1}^{K} o_k \cdot \log\ddfrac{e^{\kappa f_{k \rightarrow i} \cdot C_k}}{\sum_{\substack{C_l \in \mathcal{C}_y \\ C_l\notin \mathcal{N}_k}} e^{\kappa f_{k \rightarrow i} \cdot C_l} + \omega_\beta \sum_{\beta_n \in \mathcal{B}} e^{\kappa f_{k \rightarrow i} \cdot \beta_n} + \omega_{\Bar{y}} \sum_{C_m \in \mathcal{C}_{\Bar{y}}}e^{\kappa f_{k \rightarrow i} \cdot C_m}},
\end{equation}
where $o_k=1$ if the Gaussian is visible and $o_k=0$ otherwise.

\noindent \textbf{Updating Gaussian and Background Features.}
The Gaussian and background features $\mathcal{C}$ and $\mathcal{B}$ are updated after every gradient-update of the feature extractor. Following \cite{he2020momentum,bai2022coke}, we use momentum update for the Gaussian features:
\begin{equation}
    C_k \leftarrow C_k \cdot \sigma +  f_{k \rightarrow i} \cdot (1 - \sigma), \quad \lVert C_k \lVert = 1.
\end{equation}
The background features are simply resampled from the newest batch of training images. In particular, we remove the oldest features in $\mathcal{B}$, \ie $ \mathcal{B} = \{\beta_n\}_{n=1}^N \setminus \{\beta_n\}_{n=1}^T $.
Next, we randomly sample $T$ new background features $f_b$ from the feature map, where $f_b$ is a feature that no Gaussian contributes to and add them into the background feature set $\mathcal{B}$ (\ie $\mathcal{B} \gets \mathcal{B} \cup  \{f_b\}$). We note that $\sigma$ and $T$ are hyper-parameters of our model.
\begin{figure}
    \centering
    \includegraphics[width=.9\linewidth]{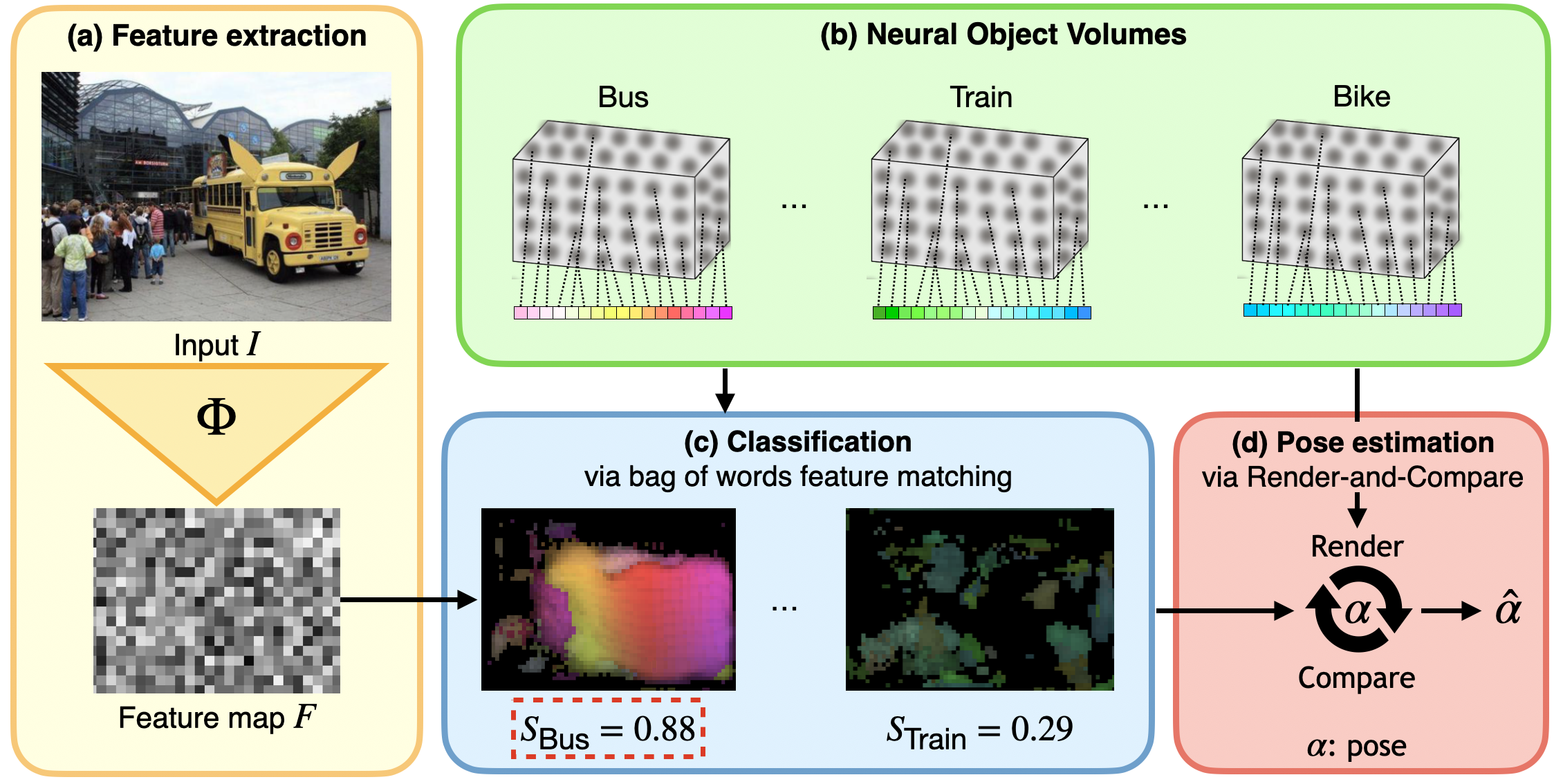}
    \caption{Overview of the classification \textit{inference pipeline}. \OURS is composed of a backbone $\Phi$ and a set of neural object volumes represented as 3D Gaussians on a cuboid shape (green box) with colored associated features. During inference, an image is first processed by the backbone into a feature map $F$. The object class is predicted by \textit{independently} matching the Gaussian features to the feature map (blue box). We color-code the detected Gaussians to highlight the interpretability of our method. Brightness shows the prediction confidence. Note that the model is only confident with the correct class even though the bus is an out-of-distribution sample. The 3D object pose can also be inferred via inverse rendering of the neural object volume (red box).}
    \label{fig:pipeline}
\end{figure}

\subsection{Inference of object category and 3D pose}
\label{sec:method:inference}

\noindent \textbf{Object Classification via Feature Matching without Geometry.}
\label{sec:method:embeddings matching}
Our classification inference pipeline is illustrated in \Cref{fig:pipeline}. 
We perform classification in a fast and robust manner via matching the extracted features to the learned Gaussian features and background features. In short, for each object category $y$, we compute the foreground likelihood $P(f_i | \mathcal{C}_y )$ and the background likelihood $P(f_i|\mathcal{B})$ on all locations $i$ in the feature map.
In this process, we do not take into account the object geometry, which reduces the matching to a simple convolution operation, hence making it very fast. To classify an image, we compare the total likelihood scores of each class average over all locations $i$.

In detail, we define a binary valued parameter $z_{i, k}$ such that $z_{i, k} = 1$ if the feature vector $f_i$ matches best to any Gaussian feature $\{C_k\}\in\mathcal{C}_y$, and $z_{i, k} = 0$ if it matches best to a background feature. 
The object likelihood of the extracted feature map $F=\mathbf{\Phi}_w(I)$ can then be computed as:
\begin{equation}
     \prod_{f_i \in F} P(f_i | z_{i, k}, y) =  
    \prod_{f_i \in F} P(f_i | C_k) ^ {z_{i, k}}
    \prod_{f_i \in F} \max_{\beta_n \in \mathcal{B}} P(f_i | \beta_n)^{1 - z_{i, k}}.
\end{equation}
As described in \Cref{sec:method:representation}, the extracted features follow a vMF distribution. Thus the final prediction score of each object category $y$ is:
\begin{equation}
    S_y = 
    \sum_{f_i \in F} \max  \{  \max_{C_k \in \mathcal{C}_y} f_i \cdot C_k, \max_{\beta_n \in \mathcal{B}} f_i \cdot \beta_n\}.
\end{equation}
The final category prediction is $\hat{y} = \argmax_{y \in Y} \{S_y\}$. 
\Cref{fig:pipeline} (blue box) illustrates the matching process for different object classes by color coding the detected Gaussians. For the correct class, the Gaussians can be detected coherently even without taking geometry into account (as can be observed by the smooth color variation), while for wrong classes this is not the case. 
Our ability to visualize the predicted kernel correspondence demonstrates an advanced interpretability of the decision process compared to standard classifiers.

\noindent \textbf{Volume Rendering for Pose Estimation.}
\label{sec:method:render_and_compare}
Given the predicted object category $\hat{y}$, we use the Gaussian features $\mathcal{C}_{\hat{y}}$ to estimate the camera pose $\alpha$ leveraging the 3D geometrical information of the neural object volumes.
Following the vMF distribution, we optimize the pose $\alpha$ via feature reconstruction: 
\begin{equation}
\label{eq:reconstruction_loss}
\mathcal{L}(\alpha) = \sum_{f_{i} \in FG}  f_i \cdot \bm{\hat{C}}_i(\alpha) + \sum_{f_b \in BG} \max_{\beta_n \in \mathcal{B}} f_b \cdot \beta_n, 
\end{equation}
where $FG$ is the set of foreground features that are covered by the rendered neural object, \ie those features for which the aggregated volume density is bigger than a threshold $FG = \{f_i \in F, \sum_{k=1}^K \rho_k(\bm{r}_{\alpha}(t)) > \theta \}$.
$BG = F \setminus FG$ is the set of features in the background.
We estimate the pose by first finding the best initialization of the object pose $\alpha$ by computing the reconstruction loss (\Cref{eq:reconstruction_loss}) for a set of pre-defined poses. Subsequently, we start gradient-based optimization using the initial pose that achieved the lowest loss to obtain the final pose prediction $\hat{\alpha}$.

\section{Experiments} \label{sec:exp}
In this section, we evaluate \OURS in terms of generalization on in-distribution and out-of-distribution data, 3D pose estimation and interpretability.
We first discuss our experimental setup (\Cref{sec:setup}), present baselines and results for classification (\Cref{sec:exp:performance_class}) and 3D pose estimation (\Cref{sec:exp:performance_pose}). Additionally, we perform in-depth evaluations of interpretability and prediction consistency, and an ablation study (\Cref{sec:exp:comprehensive_eval}).

\subsection{Setup}\label{sec:setup}
\noindent \textbf{Datasets}.
We test \OURS's robustness using four datasets where 3D pose annotations are available: PASCAL3D+ (P3D+)\cite{xiang2014beyond}, occluded-P3D+\cite{wang2020robust}, corrupted-P3D+\cite{michaelis2019corruptions}, and Out-of-Distribution-CV (OOD-CV)\cite{zhao2023ood}. PASCAL3D+ includes 12 object categories. 
Building on the P3D+ dataset, the occluded-P3D+ dataset is a test benchmark that evaluates robustness under multiple occlusion levels. 
It simulates realistic occlusion by superimposing occluders on top of the objects with three different levels: L1: 20\%-40\%, L2: 40\%-60\%, and L3:60\%-80\%, where each level has corresponding percent of the objects and the background occluded. 
P3D+ dataset does not contain significant occlusion and is therefore referred to as occlusion level 0 (L0). 
The corrupted-P3D+ corresponds to P3D+ where we apply $12$ types of corruptions \cite{hendrycks2019benchmarking,michaelis2019corruptions} to each image of the original test images, and we choose a severity level of 4 out of 5.
The OOD-CV dataset is a benchmark that includes real-world OOD examples of 10 object categories varying in terms of 5 nuisance factors: pose, shape, context, texture, and weather. 

\noindent \textbf{Implementation Details.}
Each neural object contains approximately $K=1100$ Gaussians that are distributed uniformly on the cuboid. The shared feature extractor $\Phi$ is a ResNet50 \cite{he2016deep} model with two upsampling layers and an input shape of $640 \times 800$. All features have a dimension of $D=128$ and the size of the feature map $F$ is $1/8^{th}$ of the input size.
\OURS is trained as described in \Cref{sec:method:training}, taking around 20 hours on 4 RTX 3090 with 200 epochs. During training, we use $N=2560$ background features. For each gradient step, we use $\sigma = 0.9$ for momentum update of the Gaussian features and sample $T=5$ new background features to update $\mathcal{B}$. We set $\kappa=1$ (see Appendix \ref{sup:kappa} for more details). 
We predict the object class as described in \Cref{sec:method:embeddings matching}. The feature matching for classification takes around $\mathbf{0.01}$s per sample on 1 RTX 3090, which is comparable to state-of-the-art feed-forward classification models.
\OURS can also infer the 3D object pose via inverse rendering by first evaluating the reconstruction loss on $144$ initial poses ($12$ azimuth angles, $4$ elevation angles, $3$ in-plane rotations) and subsequently starting a gradient-based inverse rendering (\Cref{eq:reconstruction_loss}) starting with lowest feature reconstruction loss as initialization. 
The pose inference pipeline takes around $0.21$s per sample on 1 RTX 3090. We note that this inference might be further optimized \eg by caching intermediate matching results efficiently.

\noindent \textbf{Evaluation.}
We evaluate our approach on two tasks: classification and 3D pose estimation. 3D pose estimation involves predicting azimuth, elevation, and rotations of an object with respect to a camera. Following \cite{zhou2018starmap}, the pose estimation error is calculated between the predicted rotation matrix $R_{\text{pred}}$ and the ground truth rotation matrix $R_{\text{gt}}$ as $\Delta\left(R_{\text{pred}}, R_{\text{gt}}\right)=\left\|\log m\left(R_{\text{pred}}^{T} R_{\text{gt}}\right)\right\|_{F}/\sqrt{2}$. We measure accuracy using two thresholds $\frac{\pi}{18}$ and $\frac{\pi}{6}$.

\noindent \textbf{Classification baselines.} We compare the performance of our approach to four competitive baseline architectures (\ie Resnet50, Swin-T, Convnext, and ViT-b-16) for the classification task. During training, these baselines are trained with a classification and a pose estimation head. Hence, 3D information is leveraged for these classification evaluations. For each baseline, we use a  classification head for which the output is the number of classes in the dataset (\ie 12 for (occluded, corrupted)-P3D+; 10 for OOD-CV). We finetune each baselines during 200 epochs.
In order to make baselines more robust, we apply standard data augmentation (i.e., scale, translation, rotation, and flipping) for each during training. More details about baselines can be found in the appendix.

\noindent \textbf{3D-Pose estimation.} We compare the performance of our approach to five other baselines for the 3D pose estimation task. For Resnet50, Swin-T, Convnext, and ViT-b-16, we consider the pose estimation problem as a classification problem (following \cite{zhou2018starmap}) by using 42 intervals of ${\sim}8.6^\circ$ for each parameter that needs to be estimated (azimuth and elevation angle, and in-plane rotation). 
We fine-tune each baseline for 200 epochs.
Similarly to classification, we apply standard data augmentation during training.
We further evaluate against NeMo \cite{wang2021nemo} that was explicitly designed for robust 3D pose estimation. We following the publicly available code and train a NeMo model  for each class.

\subsection{Robust Object Classification} \label{sec:exp:performance_class}

We first evaluate the performance on IID data. As the L0 column of \Cref{tab:classification_table} shows, our approach achieves $99.5\%$ for classification, which is comparable to other baselines.
Furthermore, our approach manages to robustly classify images in various out-of-distribution scenarios. From \Cref{tab:classification_table}, we can see that our representation allows to outperform all other traditional baselines with around 6\% accuracy on average for different levels of occlusions and with up to 33\% accuracy boost for images under five different types of nuisances in OOD-CV. For corrupted data, our approach also outperforms the baselines on average. 
In summary, \OURS achieves \textbf{significant improvements in OOD generalization} while \textbf{maintaining state-of-the-art accuracy for IID} data for classification. Finally, it is worth noting that our approach is \textbf{more consistent} than all baselines. Our approach consistently outperforms baselines over all nuisances, which indicates the intrinsic robustness in our architecture. 

\begin{table}[ht]
    \small
    \caption{Classification accuracy results on P3D+, occluded-P3D+, OOD-CV and corrupted-P3D+ datasets. L0 corresponds to unoccluded images from Pascal3D+, and occlusion levels L1-L3 are from occluded-P3D+ dataset with occlusion ratios stated in \Cref{sec:setup}. Our approach performs similarly in IID scenarios, while steadily outperforming all baselines in OOD scenarios. First is highlighted in \textbf{bold}, second is \ul{underlined}. Higher is better. Full results in Appendix.}
  \begin{center}
  \tabcolsep=0.30cm
  \resizebox{\textwidth}{!}{%
    \begin{tabular}{l|c|cccc|c|c}
    \toprule
    Dataset & P3D+ &\multicolumn{4}{c|}{occluded-P3D+} & OOD-CV & corrupted-P3D \\
    \hline
    Nuisance & L0 & L1 & L2 & L3 & Mean & Mean & Mean \\ 
    \midrule
    Resnet50 & 99.3 & 93.8 & 77.8 & 45.2 & 72.3 & 51.4&78.7\\
          
    Swin-T & \ul{99.4} & 93.6 & 77.5 & 46.2 & 72.4 & \ul{64.2}&78.9\\
    
    Convnext & \ul{99.4} & \ul{95.3} & \ul{81.3} & \ul{50.9} & \ul{75.8} & 56.0&85.6\\

    ViT-b-16 &99.3 & 94.7 & 80.3 & 49.4 & 74.8  & 59.0&\ul{87.6}\\
    \OURS & \textbf{99.5} & \textbf{97.2}  & \textbf{88.3}  & \textbf{59.2}  & \textbf{81.6}   & \textbf{85.2} & \textbf{91.3}\\
    \bottomrule
    \end{tabular}}
     \end{center}
     \label{tab:classification_table}
     \end{table}
     \raggedbottom

\begin{table}[ht]
\caption{3D-Pose Estimation results for different datasets. A prediction is considered correct if the angular error is lower than threshold (\ie $\frac{\pi}{6}$, and $\frac{\pi}{18}$). Higher is better. Our approach shows it is capable of robust 3D pose estimation that performs similarly to the current state-of-the-art. Note that models marked with a `*' possess an explicit 3D representation.}
 \begin{center}
    \tabcolsep=0.11cm
    \small
        \begin{tabular}{l|cccc|cccc}
    \toprule
    Dataset & P3D+ &\makecell{occ-\\P3D+} & \makecell{cor-\\P3D+} & OOD-CV & P3D+ &\makecell{occ-\\P3D+} & \makecell{cor-\\P3D+} & OOD-CV \\
    \hline
    Threshold & \multicolumn{4}{c|}{$\pi/6$} & \multicolumn{4}{c}{$\pi/18$} \\
    \midrule
    Resnet50 & 82.2 & 53.8& 33.9& \ul{51.8}& 39.0 & 15.8 & 15.8 & 18.0\\
    Swin-T & 81.4 & 48.2& 34.5& 50.9& 46.2 & 16.6 & 15.6 & 19.8 \\
    Convnext & 82.4 & 49.3 & 37.1& 50.7& 38.9 & 14.1 & 24.1 & 19.9 \\
    ViT-b-16 & 82.0 & 50.8& 38.0& 48.0& 38.0 & 15.0 & 21.3 & 21.5 \\          
    NeMo* & \ul{86.1} & \ul{62.2}& \ul{48.0}& 51.6& \textbf{61.0} & \ul{31.8} & \textbf{43.4} & \ul{21.9}\\
    
    \OURS* & \textbf{88.2} & \textbf{63.1}& \textbf{49.1}& \textbf{52.9}& \ul{59.1} & \textbf{32.7}& \textbf{43.4} & \textbf{22.8}\\ 
    \bottomrule

    \end{tabular}
    \label{tab:pose_table}
  \end{center}
\end{table}
\raggedbottom

\subsection{Robust 3D Pose Estimation}
\label{sec:exp:performance_pose}

According to the results in \Cref{tab:pose_table}, our approach outperforms all baselines significantly across IID and OOD scenarios, while also exceeding the performance of NeMo \cite{wang2021nemo}, the current state-of-the-art method for robust 3D pose estimation. This higher performance can be attributed to the better optimization of volume rendering which gives more stable gradients compared to mesh-based differentiable rendering.

\subsection{Comprehensive assessment of our representation}
\label{sec:exp:comprehensive_eval}

\begin{figure}
\centering
            \begin{subfigure} {.22\textwidth}
                \centering
                \includegraphics[width=\linewidth]{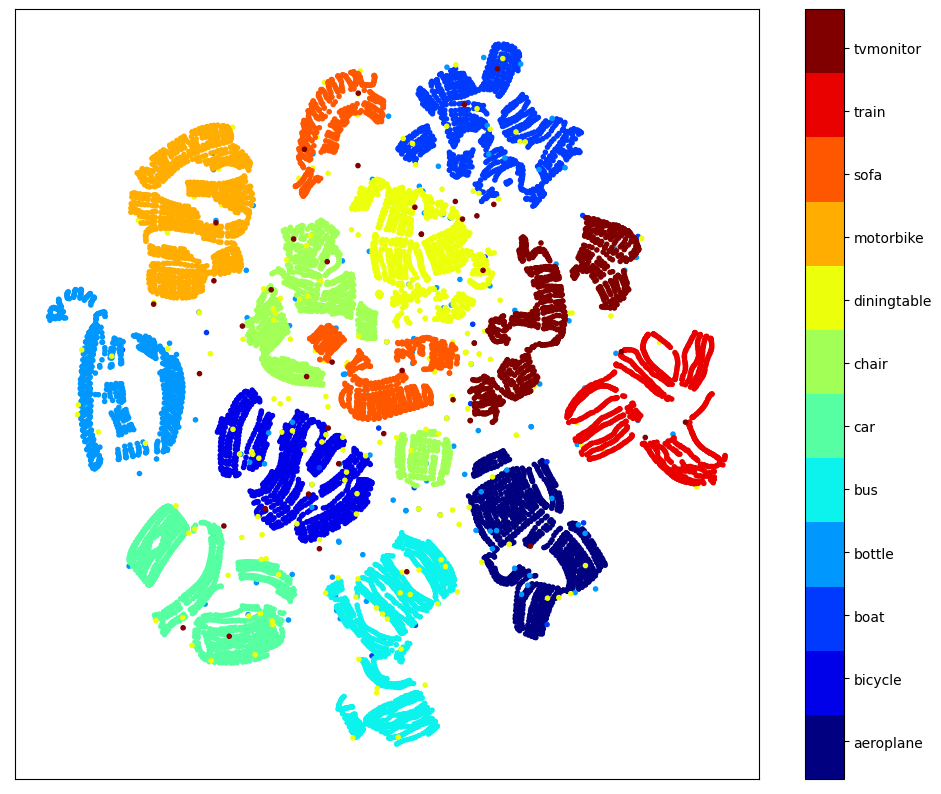}
                \caption{\textbf{\OURS}: t-SNE plot of learned feat. $\mathcal{C}$.}
                \label{fig:tsne_classes}
            \end{subfigure}
            \hspace{.01\textwidth}
            \begin{subfigure} {.22\textwidth}
                \centering
                \includegraphics[width=\linewidth]{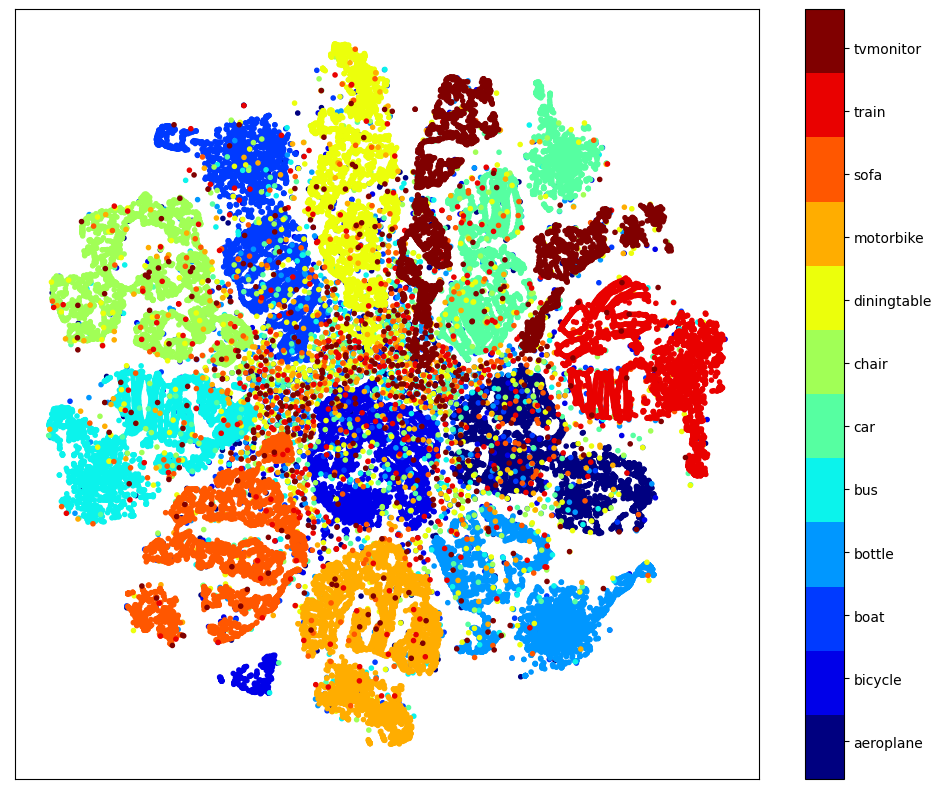}
                \caption{Nemo: t-SNE plot of learned features $\Theta$.}
                \label{fig:tsne_classes_nemo}
            \end{subfigure}
            \hspace{.01\textwidth}
            \begin{subfigure} {.22\textwidth}
                \centering
                \includegraphics[width=\linewidth]{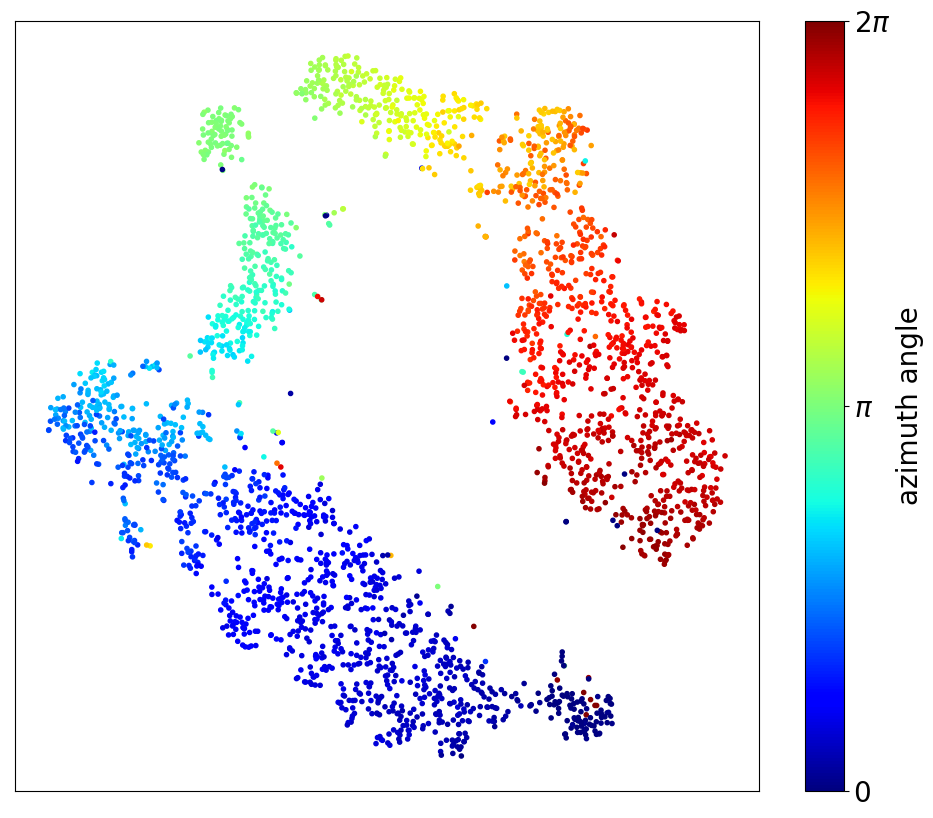}
                \caption{\textbf{\OURS}: t-SNE plot of all car samples.}
                \label{fig:tsne_pose}
            \end{subfigure}
            \hspace{.01\textwidth}
            \begin{subfigure} {.22\textwidth}
                \centering
                \includegraphics[width=\linewidth]{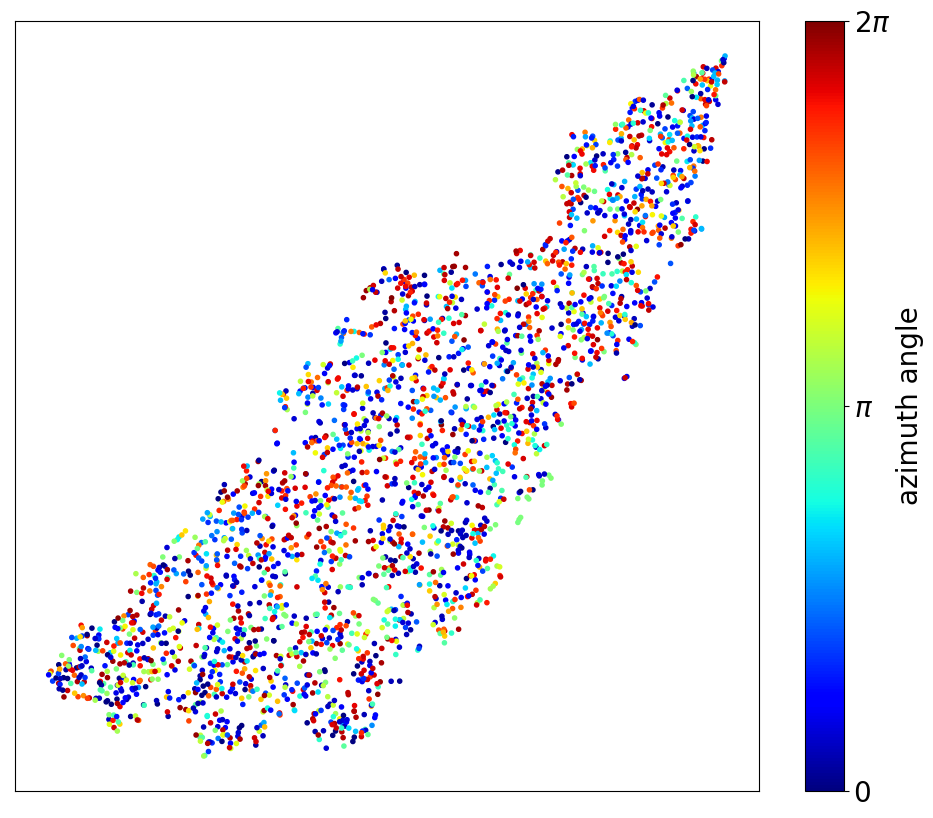}
                \caption{Resnet50: t-SNE plot of all car samples.}
                \label{fig:tsne_pose_resnet}
            \end{subfigure}
            \caption{\textbf{(a-b)} t-SNE plots comparing (a) the learned features $\mathcal{C}$ of our approach and (b) the learned vertex features $\Theta$ of NeMo. As can be seen, our contrastive loss allows a much clearer distribution of the space while keeping Gaussian features from different classes far from each other (while the low-quality clustering observed in (b) may likely originates from the ImageNet pretraining). \textbf{(c-d)} t-SNE plots of the mean extracted feature for each car image of the test dataset. We observe a very clear organization of the samples according to the azimuth angle for (c) our approach while this organization is completely absent in (d) other feed-forward baselines (\eg Resnet50).}
            \label{fig:interpretability_tsne}
\end{figure}

\begin{figure}
\centering
    \centering
    \includegraphics[width=0.9\linewidth]{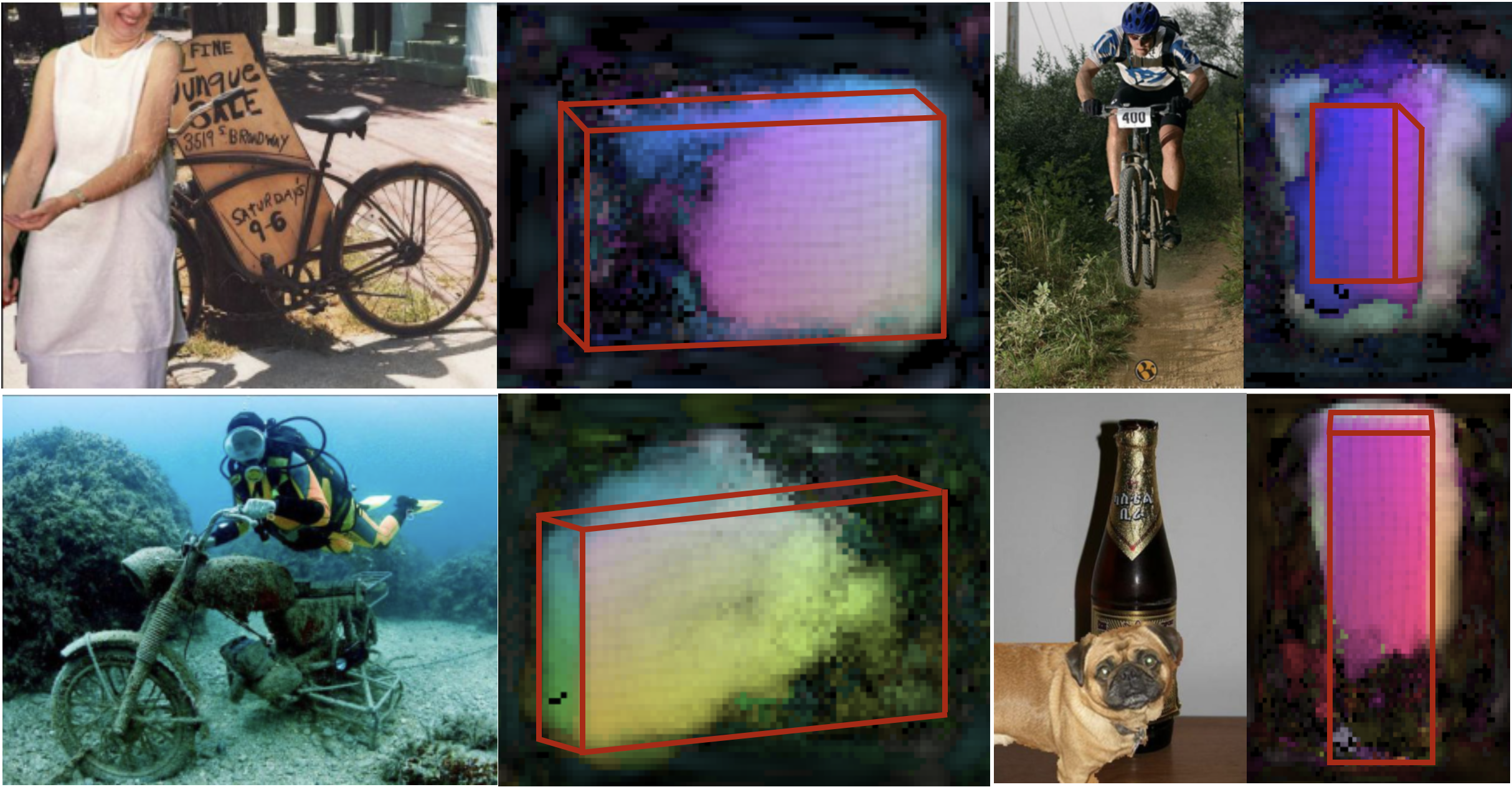}
    \label{fig:quali1}
    \caption{Four qualitative results that were misclassified by ViT-b-16. We show for each: (left) the input image and (right) the extracted feature map and the predicted 3D pose overlaid. 
    We color coded the features by encoding the color as a function of $\mu_k$ of the matched Gaussian $C_k$ (as done in NOCS \cite{NOCS}). Hence, a smooth color gradient shows a high quality matching. In the extracted features, the brightness illustrates the confidence of matching with the Gaussian features.}
    \label{fig:qualitative}
\end{figure}

\noindent \textbf{Interpretability.}
Our explicit volumetric representation can be leveraged to predict the object class and its pose in an image. Insightful information also lies in the Gaussian feature matching between image features and the neural object volumes. Visualizing the matching results (see \Cref{fig:qualitative} and more results including videos in Appendix \ref{sec:supp:visualizations} and \Cref{fig:supp:qualitative}), reveals which object parts are perceived by the model at any given location in the image. As illustrated in \Cref{fig:qualitative}, the occluded parts of the objects are not colored, meaning that no Gaussian has been matched with these image features.
\Cref{fig:tsne_classes,fig:tsne_pose} moreover show that our learned features are disentangled and encode useful information in terms of object classes and 3D pose. When comparing \Cref{fig:tsne_classes,fig:tsne_classes_nemo}, we observe different cluster for each category while the low-quality clustering observed in the latter Figure may likely originates from the ImageNet pretraining. When comparing \Cref{fig:tsne_pose,fig:tsne_pose_resnet}, we observe a consistent distribution of car instances depending of their pose in the former while in the latter, features don't have any explicit organization in terms of pose.

\begin{table}
\centering
\begin{minipage}{.45\linewidth}
  \centering
    \caption{Consistency of predictions for classification and 3D pose estimation. The last line shows the accuracy of those images with both object pose (threshold: $\frac{\pi}{6}$) and class correctly predicted. "cls." stands for classification.}
\label{tab:exp:consistency}
        \begin{tabular}{l|c|ccc}  
        \toprule
        Dataset& P3D+ &\multicolumn{3}{c}{occ-P3D+} \\
        \hline
         Task& L0 & L1 & L2 & L3 \\
        \midrule
        classification & 99.5 &97.2 &88.3 &59.2\\
        3D pose (th: $\frac{\pi}{6}$) & 90.1 &82.6&71.3&52.7\\
        cls. \& 3D pose & 89.8 &81.9&70.2&50.2\\
        \bottomrule
    \end{tabular}
\end{minipage}%
\hspace{0.06\linewidth}
\begin{minipage}{.45\linewidth}
  \centering
    \caption{Ablation studies of the background features and the shape of the representation on P3D+ and occluded-P3D+. We ablate both the background features and the object shape.}
    \label{tab:ablations}
    \begin{tabular}{cc|c|ccc}
        \toprule
        \multicolumn{2}{c|}{Components}& P3D+ &\multicolumn{3}{c}{occ-P3D+} \\
        \hline
         $\mathcal{B}$ & Shape& L0 & L1 & L2 & L3 \\ 
        \midrule
                          & single feature  & 93.2  & 90.3  & 80.4  & 44.0  \\
                         $\checkmark$ & sphere  & 99.3  & 97.0  & 87.9  & 59.0  \\
                                     & cuboid  & 99.3  & 97.0  & 85.7  & 53.0  \\
          $\checkmark$ & cuboid  & \textbf{99.5} & \textbf{97.2}  & \textbf{88.3}  & \textbf{59.2}  \\
        \bottomrule
    \end{tabular}
\end{minipage}
\end{table}

\noindent \textbf{Consistency.}
Another valuable characteristics of our approach lies in the fact that it is consistent between the different tasks. In \Cref{tab:exp:consistency}, we observe that the accuracy for samples that have correct class and pose prediction are limited by the pose estimation itself since they are fairly similar. In IID scenarios, the difference is only of $0.3\%$, while in OOD scenarios the difference is around $1\%$ on average.
We believe that this consistency comes from the common explicit volumetric representation that is shared for all tasks. Such a behavior would not be expected between different task-specific models that are trained separately.

\noindent \textbf{Efficiency.} 
For classification, \OURS matches real-time performance as other CNN or transformer-based baselines, reaching an inference speed of 50FPS. Despite variations in parameter numbers among baselines (Swin-T: 28M, ViT-b-16: 86M, \OURS: 83M), we find no correlation of the parameter count with OOD robustness. 
For pose estimation, compared to the render-and-compare-based method, NeMo \cite{wang2021nemo}, our model uses a significantly lower parameter count (\OURS: 83M, NeMo: 996M, \ie an impressive 12x reduction) due to our class-agnostic backbone. We also observe that our model converges faster in the render-and-compare optimization compared to NeMo (\OURS: 30 steps, NeMo: 300 steps), which can be attributed to our class-contrastive representation and the neural object volumes which yields better gradient during the optimization. More detailed comparisons can be found in Appendix \ref{sup:sec:efficiency}.

\noindent \textbf{Ablations.}
We ablate the geometry selection and background features during training. In \OURS, we form a cuboid with Gaussians. An alternative could be to choose to (1) employ a finer-grained representation, (2) utilize a single Gaussian to represent the volume, or (3) adopt a spherical geometric representation instead of a cuboid. Opting for the first alternative would necessitate either establishing a deformable geometry or treating each sub-category as a distinct class, which is beyond the scope of this work.
In \Cref{tab:ablations} (more details in Appendix \ref{sec:supp:ablation:sphere}), we ablate the shape of our 3D representation. As expected, the choice of the representation's shape does not have a pronounced influence on performance, as was already underlined for meshes \cite{wang2021nemo}. However, we observe a slight advantage in favor of the cuboid shape which may be attributed to the cuboid's closer approximation of the true shape. Additionally, we considered a \textit{single feature vector} approach, where a single Gaussian per class is employed during training using contrastive learning. We observe a performance drop by up to $15\%$, which highlights the importance of the geometry during training. 
These findings corroborate the classification results of our method: by selectively omitting some geometric information (\ie the 3D structure), we can reach similar outcomes while significantly enhancing computational efficiency. Ultimately, we note that the background model $\mathcal{B}$ is beneficial during training since it promotes greater dispersion among neural features. This proves to be useful for inference in cases marked by occlusions but does not have visible effect in IID scenarios.

\section{Limitations and Future Work}
Despite the strong generalization performance, multi-task capabilities and interpretability, the proposed \OURS model also has limitations. 
For example, the current necessity for pose annotation during training is suboptimal.
However, we maintain an optimistic outlook regarding future advancements that may mitigate this requirement. 
Notably, recent studies \cite{goodwin2022zero,goodwin2023you,Yang20tro-teaser} are investigating the demanding task of aligning objects with minimal examples, indicating a promising research trajectory.
Currently, the geometry of the neural object volumes is fixed, which is suboptimal for object categories with large intra-class variability. 
Moving forward, exploring methods for flexible deformations would likely enhance the performance of the model, while also enabling the accurate segmentation of objects, thus enriching the scope of the proposed framework.

\section{Conclusion}
In this work, we introduced \OURS, an architecture for object classification with an explicit compositional 3D object representation.
We presented a framework for learning neural object volumes with associated features across various categories using annotated 3D object poses.
Our experimental results shows that \OURS achieves much higher robustness compared to other state-of-the-art architectures under OOD scenarios, \eg occlusion and corruption, with competitive performance in in-distribution scenarios, and similar inference speed for image classification. 
Further, we demonstrate \OURS can also estimate 3D pose accurately by following an inverse rendering approach, achieves an enhanced human interpretability and consistency of multi-task predictions.
In conclusion, our results showcase \OURS as a pioneering step towards highly advanced generalization capabilities in vision models.

\section*{Acknowledgements}
Adam Kortylewski acknowledges support via his Emmy Noether Research Group funded by the German Reserach Foundation (DFG) under Grant No. 468670075. Alan Yuille acknowledges support from Army Research Laboratory award W911NF2320008 and Office of Naval Research N00014-21-1-2812.

\bibliographystyle{splncs04}
\bibliography{main}

\newpage
\beginsupplement
\section*{Supplementary material}
We provide additional results and discussions to support the experimental results in the main paper.

\section{NeMo baseline extension}
\label{sup:sec:nemo_extension}
In the following, we introduce in more details how Nemo \cite{wang2021nemo} can be na\"{i}vely extended to perform classification and how does it perform for this new task.

\subsection{Extension procedure}
NeMo \cite{wang2021nemo} is originally designed to perform 3D-pose estimation and the class of the object is considered as known. We leverage one class-specific trained NeMo model for each object category (12 for Pascal3D+; 10 for OOD-CV). We use the exact same procedure as in the original paper\cite{wang2021nemo} for training the class-specific NeMo models and for inference. Since NeMo relies on a render-and-compare approach, we can obtain the reconstruction loss from the final prediction for each candidate class which we leverage to assess the quality of the predicted 3D-pose. Finally, the class corresponding to the lowest loss corresponds to the predicted class.

\subsection{Classification}

In this section, we compare results between the previously described na\"{i}ve extension of NeMo (\ie ext.-NeMo) and our approach for classification. Table \ref{tab:classification_table_nemo_comparison} shows that for classification, our approach considerably outperforms the na\"{i}ve extension of NeMo for Pascal3D+ and OOD-CV. In the most challenging scenarios (e.g., L3 occlusion and weather), our approach even performs more than 2x better. Besides this substantial performance improvement, the computation requirements are notably lower (both in terms of memory and temporal requirements). Hence, this demonstrates that learning the neural textures in a discriminative manner between classes is crucial to observe good classification performances. 

\begin{table*}[h]
\small
\caption{Classification accuracy results on PASCAL3D+, occluded-PASCAL3D+ and OOD-CV datasets. Higher is better. We observe a significant performance improvement between the na\"{i}ve NeMo extension (referred to as "ext.-Nemo") and our approach. Our approach performs better in all scenarios. Abbreviations: "cont." stands for context, "text." stands for texture, and "weat." stands for weather.}
  \begin{center}
  \tabcolsep=0.11cm
    
    \begin{tabular}{l|c|cccc|cccccc}
    \toprule
    Dataset & P3D+ &\multicolumn{4}{c|}{occluded-P3D+} & \multicolumn{6}{c}{OOD-CV} \\
    \hline
    Nuisance & L0 & L1 & L2 & L3 & mean & cont. & pose & shape & text. & weat. & mean \\ 
    
    \midrule
    ext.-NeMo & 88.0 & 72.5 & 49.3 & 22.3 & 48.0 & 52.2 & 43.2 & 54.8 &45.5 & 40.4 & 46.3\\
    \OURS & \textbf{99.5} & \textbf{97.2}  & \textbf{88.3}  & \textbf{59.2}  & \textbf{81.6}  & \textbf{85.3}  & \textbf{88.1}  & \textbf{83.6}  & \textbf{90.1} & \textbf{82.8} & \textbf{85.2} \\
    \bottomrule
   
    \end{tabular}
    \label{tab:classification_table_nemo_comparison}
  \end{center}
\end{table*} 

\subsection{3D Pose Estimation}
3D Pose estimation is not affected by any changes of the extension procedure. Hence, results reported in the main paper for NeMo still stand for the extended version of NeMo.

\section{Additional results}
In this section, we supplement the findings outlined in the main paper and show results for $\frac{\pi}{18}$ and $\frac{\pi}{6}$ thresholds. Then we show the full results for all tasks for all corruptions of the corrupted-PASCAL3D+ dataset.

\subsection{Classification}
Tables \ref{sup:tab:classification_table1} and \ref{tab:sup_corupt_results} show complementary results to the ones shown in the paper. We observe similar trends for all metrics than the ones presented in Sections \ref{tab:classification_table}. Our approach shows much higher performances compared to current state-of-the-art (SOTA) for the classification task.
\begin{table}[h]
    \small
    \caption{Classification accuracy results on PASCAL3D+, occluded-PASCAL3D+ and OOD-CV  datasets. First is highlighted in \textbf{bold}, second is \ul{underlined}. L0 corresponds to unoccluded images from Pascal3D+, and occlusion levels L1-L3 are from occluded-PASCAL3D+ dataset with occlusion ratios stated in \ref{sec:setup}. \OURS performs similarly in IID scenarios, while steadily outperforming all baselines in OOD scenarios. Abbreviations: "cont." stands for context, "text." stands for texture, and "weat." stands for weather.}
    \label{sup:tab:classification_table1}
  \begin{center}
  \tabcolsep=0.11cm
    \begin{tabular}{l|c|cccc|cccccc}
    \toprule
    Dataset & P3D+ &\multicolumn{4}{c|}{occluded-P3D+} & \multicolumn{6}{c}{OOD-CV}  \\
    \hline
    Nuisance & L0 & L1 & L2 & L3 & mean & cont. & pose & shape & text. & weat. & mean \\ 
    
    \midrule
    Resnet50 & 99.3 & 93.8 & 77.8 & 45.2 & 72.3 & 45.1 & 61.2 & 55.2 & 48.3 & 47.3 & 51.4\\
          
    Swin-T & \ul{99.4} & 93.6 & 77.5 & 46.2 & 72.4 & \ul{63.0} & 71.4 & \ul{65.9} & \ul{61.4} & \ul{59.6} & \ul{64.2}\\
    
    Convnext & \ul{99.4} & \ul{95.3} & \ul{81.3} & \ul{50.9} & \ul{75.8}& 53.6 & 61.2 & 60.8 & 57.2 & 47.1 & 56.0\\

    ViT-b-16 &99.3 & 94.7 & 80.3 & 49.4 & 74.8 & 57.8 & \ul{67.3} & 61.0 &54.7 & 54.5 & 59.0\\

    \OURS & \textbf{99.5} & \textbf{97.2}  & \textbf{88.3}  & \textbf{59.2}  & \textbf{81.6}  & \textbf{85.3}  & \textbf{88.1}  & \textbf{83.6}  & \textbf{90.1} & \textbf{82.8} & \textbf{85.2} \\

    \bottomrule
    \end{tabular} 
     \end{center}
     \end{table}

\subsection{Additional results for $\frac{\pi}{18}$ and $\frac{\pi}{6}$ thresholds}
Table \ref{tab:sup_pose_table_mederr} shows complementary results to the ones shown in the paper. We observe similar trends for all metrics than the ones presented in Sections \ref{sec:exp:performance_pose}. Our approach shows equivalent performances to current state-of-the-art (SOTA) for the 3D-pose estimation task even though it is not specifically designed to perform this task.
\begin{table*}[h]
\small
\caption{Pose Estimation results on (occluded)-PASCAL3D+, and OOD-CV dataset. Pose accuracy is evaluated for error under two thresholds: $\frac{\pi}{6}$ and $\frac{\pi}{18}$ separately. Noticeably, our approach has equivalent performances to current SOTA for 3D-pose estimation event though it has not been specifically designed for this task. Abbreviations: "cont." stands for context, "text." stands for texture, and "weat." stands for weather.}
 \begin{center}
  \tabcolsep=0.11cm
    
    \begin{tabular}{l|c|cccc|cccccc}
    
    \toprule
    \multicolumn{11}{c}{Threshold: $\frac{\pi}{18}$}\\
    \midrule
    Dataset & P3D+ & \multicolumn{4}{c|}{occluded-P3D+} & \multicolumn{6}{c}{OOD-CV} \\
    \hline

    Nuisance & L0 & L1 & L2 & L3 & mean & cont. & pose & shape & text. & weat. & mean \\
    \midrule
    Resnet50 & 33.8 & 22.4 & 15.8 & 9.1 & 15.8 & 15.5 & 12.6 & 15.7 & 22.3 & 23.4 & 18.0 \\

    Swin-T & 29.7 & 23.3 & 15.6 & 10.8 & 16.6 & 18.3 & 14.4 & 16.9 & 21.1 & 26.3 & 19.8 \\
    
    Convnext & 38.9 & 22.8 & 12.8 & 6.6 & 14.1 & 18.1 & \textbf{14.5} & 16.5 & 21.7 & 26.6 & 19.9 \\

    ViT-b-16 & 38.0 & 23.9 & 13.7 & 7.4 & 15.0 & 24.7 & 13.8 & 15.6 & 25.0 & 28.3 & 21.5 \\
 
    NeMo & 61.0 & 45.0 & 30.7 & 14.6 & 31.8 & 21.9 & 6.9 & 19.5 & \textbf{34.0} & 30.4 & 21.9\\
           
    \OURS & \textbf{69.5} & \textbf{47.5} & \textbf{31.6} & \textbf{16.2} & \textbf{32.7} & \textbf{26.2} & 12.1 & \textbf{25.8} & 32.5 & \textbf{34.5} & \textbf{22.8}\\

    \end{tabular}
    \begin{tabular}{l|c|cccc|cccccc}
    \toprule
    \multicolumn{11}{c}{Threshold: $\frac{\pi}{6}$}\\
    \midrule
    Dataset & P3D+ & \multicolumn{4}{c|}{occluded-P3D+} & \multicolumn{6}{c}{OOD-CV} \\
    \hline

    Nuisance & L0 & L1 & L2 & L3 & mean & cont. & pose & shape & text. & weat. & mean \\
    \midrule
    
     Resnet50 & 82.2 & 66.1 & 42.1 & 53.8 & 57.8 & 34.5 & 50.5 &53.1& \textbf{61.5} & 60.0 & 51.8 \\

    Swin-T & 81.4 & 58.5 & 47.3 & 38.8 & 48.2 & 52.3 & 41.1 & 45.7 & 50.1 & 64.9 & 50.9 \\
    
    Convnext & 82.4 & 63.7 & 47.9 & 36.4 & 49.3 & 51.7 & \textbf{43.4} & 44.8 & 48.0 & \textbf{65.9} & 50.7  \\

    ViT-b-16 & 82.0 & 65.4 & 49.5 & 37.6 & 50.8 & 54.7 & 34.0 & 49.5 & 59.1 & 59.0 & 51.3 \\
 
    NeMo & 86.1 & 75.9 & 63.9 & 45.6 & 62.2 & 50.3 & 35.3 & 49.6 & 57.5 & 52.2 & 51.6\\
       
    \OURS & \textbf{90.1} & \textbf{82.6} & \textbf{71.3} & \textbf{52.7} & \textbf{63.1} & \textbf{54.9} & 24.8 & \textbf{54.6} & \textbf{61.5} & 55.3 & \textbf{52.9}\\
    
    \bottomrule
    
    \end{tabular}
    \label{tab:sup_pose_table_mederr}
  \end{center}
  
 \end{table*}
\subsection{Full corrupted-PASCAL3D+ results}
Table \ref{tab:sup_corupt_results} shows corrupted-PASCAL3D+ results for classification and for all corruptions. Similarly to what have been discussed previously, \OURS significantly outperforms all baselines.

\begin{landscape}
\begin{table*}
    \caption{Classification results for (corrupted)-PASCAL3D+. 3D-pose estimation is evaluated for error under two thresholds: $\frac{\pi}{6}$ and $\frac{\pi}{18}$ separately. Our approach significantly outperforms all baselines. Abbreviations: "bright." stands for brightness.}
  \begin{center}
  \tabcolsep=0.11cm
    \begin{tabular}{c|l|c|ccccccccccccc} 
    \toprule
    \multicolumn{2}{c|}{Dataset} & P3D+ & \multicolumn{13}{c}{corrupted-P3D+} \\
    \hline
    \multicolumn{2}{c|}{Nuisance} & L0 & \makecell{defocus\\ blur} & \makecell{glass\\ blur} & \makecell{motion\\ blur} & \makecell{zoom\\ blur} & snow & frost & fog & \makecell{bright.} & contrast & \makecell{elastic\\ transform} & pixelate & jpeg & mean\\ 

    \midrule
    \multirow{6}{*}{\makecell{classification:\\ $ACC \uparrow$}} 
    &Resnet50 & 99.3 & 67.6 & 41.4 & 73.5 & 87.5 & 84.4 & 84.3 & 93.9 & 98.0 & 90.0 & 46.4 & 82.1 & 95.5 & 78.7\\
          
    &Swin-T & \ul{99.4} & 60.7 & 37.1 & 70.9 & 81.3 & 88.5 & 91.6 & 95.4 & 97.9 & 92.1 & 56.3 & 79.2 & 95.3 & 78.9\\
    &Convnext & \ul{99.4} & \ul{70.1} & 58.7 & 76.5 & \textbf{90.0} & \textbf{92.3} & \ul{92.9} & \textbf{98.5} & \textbf{99.2} & \textbf{98.4} & 67.6 & 84.2 & \textbf{98.7} & 85.6\\

    &ViT-b-16 & 99.3 & 64.5 & \textbf{78.1} & \ul{80.3} & \ul{88.2} & \ul{91.2} & \textbf{94.1} & 90.5 & \ul{98.7} & 85.1 & \ul{84.8} & \ul{96.9} & \textbf{98.7} & \ul{87.6}\\
    &\OURS &\textbf{99.5}&\textbf{90.5}&\ul{65.7}&\textbf{86.4}&{84.2}&\ul{91.2}&{89.5}&\ul{98.4}&\ul{98.4}&\ul{97.1}&\textbf{97.2}&\textbf{97.1}&\ul{98.4}& \textbf{91.3} \\

    \bottomrule

    \end{tabular}
    \label{tab:sup_corupt_results}
  \end{center}
\end{table*}
\end{landscape}

\subsection{3D-pose initialization results}
To initiate the render-and-compare process, we require an initial pose denoted as $\alpha_{init}$. We achieve this by pre-sampling 144 distinct feature maps and subsequently calculating the similarity between these extracted features from the image and the pre-rendered maps. The initial pose $\alpha_{init}$ is then determined as the pose corresponding to the highest similarity between the rendered map and the image feature map.

Importantly, by utilizing $\alpha_{init}$ as a coarse 3D pose prediction, we can achieve a remarkable computation speed of approximately 0.04 seconds per sample on a single RTX 3090 GPU. This represents an 80\% reduction in computation time compared to the full pipeline of our approach. We oserve in Table \ref{supp:tab:pose_table} that $\frac{\pi}{6}$ results are consistent with the full pipeline. However, the $\frac{\pi}{18}$ results suffer from the coarse prediction importantly. The coarse prediction performs quite well in terms of $\frac{\pi}{6}$ but drops significantly compared to our full pipeline.

\begin{table}[h]
\small
\caption{3D-Pose Estimation results for different datasets. A prediction is considered correct if the angular error is lower than a given threshold (\ie $\frac{\pi}{6}$ and $\frac{\pi}{18}$). The coarse pose initialization is quite accurate (\ie around $2\%$ drop in performance for P3D+) when looking at the coarse metric of accuracy below $\frac{\pi}{6}$ but is significantly lower when looking at the latter metric (\ie $\frac{\pi}{18}$). Abbreviations: "Pose init." stands for Pose initialization.}
 \begin{center}
    \tabcolsep=0.11cm
        \begin{tabular}{l|cccc|cccc}
    \toprule
    Dataset & P3D+ &\makecell{occ-\\P3D+} & \makecell{cor-\\P3D+} & OOD-CV & P3D+ &\makecell{occ-\\P3D+} & \makecell{cor-\\P3D+} & OOD-CV \\
    \hline
    Threshold & \multicolumn{4}{c|}{$\pi/6$} & \multicolumn{4}{c}{$\pi/18$} \\
    \midrule
    Pose init. & 84.3 & 52.8& 34.2& 43.7&29.8 &17.0&21.1 & 18.8\\ 
    \OURS & \textbf{88.2} & \textbf{63.1}& \textbf{49.1}& \textbf{52.9}& \textbf{59.1} & \textbf{32.7}& \textbf{43.4} & \textbf{22.8}\\ 
    \bottomrule

    \end{tabular}
    \label{supp:tab:pose_table}
  \end{center}
\end{table}

\subsection{Efficiency}
\label{sup:sec:efficiency}

For classification, our method attains real-time performance comparable to other CNN or transformer-based baselines, consistently handling over 50FPS, as detailed in Table \ref{sup:tab:overview}. Despite variations in parameter numbers among these baselines, we find no discernible correlation between parameter count and OOD robustness. Notably, Table \ref{sup:tab:overview} illustrates that our approach outperforms all baselines significantly at OOD generalization, even with similar inference speeds.

In the realm of pose estimation, our model demonstrates a substantial reduction in parameter count when compared to the render-and-compare-based method NeMo \cite{wang2021nemo}, as indicated in Table \ref{sup:tab:overview}. This reduction is mostly due to our class-agnostic backbone. Additionally, our inference speed surpasses NeMo by approximately 2 times. This acceleration is primarily due to the fewer steps involved in our render-and-compare process (Ours: 30 steps, NeMo: 300 steps), driven by the observed quicker convergence towards local optima facilitated by our class-contrastive representation and better gradient during the optimization. Although CNN and transformer-based baselines exhibit higher inference speeds for pose estimation, their performance in OOD scenarios is notably lower.

In terms of floating-point operations (FLOPs), Table \ref{sup:tab:overview} reveals a stark contrast, with NeMo requiring a significantly higher number of FLOPs at 3619 GFLOPs, whereas our approach demands only 301 GFLOPs. The slightly increased number of operations in our approach in comparison to other CNN and transformer-based methods can be mainly attributed to the final feature matching procedure.

\begin{table*}[h]
\small
\caption{Overview of the parameter counts and computation time and cost for each method as well as performances in OOD scenarios (\ie mean over all nuisances of the OOD-CV dataset). NeMo refers to \cite{wang2021nemo}. For our approach, we show differnet values for the GFlops, for the classification and pose estimation pipeline, respectively. Abbreviations: "class." stands for classification, "pose est." stands for pose estimation.}
  \begin{center}
  \tabcolsep=0.11cm
    \begin{tabular}{l|cccccc}
    \toprule
    Method &\#Parameter&GFlops&\makecell{Inference\\class. (s)}&\makecell{Inference\\pose est. (s)}&\makecell{Accuracy\\class.}&\makecell{Accuracy\\pose est.}\\
    \hline
    Resnet50 & 84M&84.5&0.01 &0.01&51.4&51.8\\
    Swin-T & 28M&60.8&0.01 &0.01&64.2&50.9\\
    Convnext & 30M&91.1&0.01 & 0.01&56.0&50.7\\
    ViT-b-16 & 86M&22.6&0.01 & 0.01&59.0&48.0\\
    NeMo & 996M & 3619.2& NA & 0.41& NA & 48.0\\ 
    \OURS & 83M& 98.0 - 301.6& 0.02 & 0.21 & \textbf{85.2} & \textbf{52.9}\\
    \bottomrule
   
    \end{tabular}
    \label{sup:tab:overview}
  \end{center}
\end{table*}

\subsection{Ablation study}
\label{sec:supp:ablation}
In the main paper, we conducted ablation experiments exclusively on two datasets, namely Pascal3D+ and occluded-P3D+. This choice was necessitated by computational resource constraints. It is important to note that we do not possess any compelling rationale to believe that the outcomes observed in the aforementioned datasets would significantly differ had we examined the remaining two datasets.

In the main paper, we only evaluated ablations for two datasets (\ie Pascal3D+ and occluded-P3D+) due to computation resources. We do not have any reason to think that findings made on the aformentioned datasets would be any different if studied on the two remaining datasets.

\subsubsection{Shape ablation}\label{sec:supp:ablation:sphere}
In order to evaluate our approach with a different shape, we studied in more details the performances using a spherical shape. We used the same  neural object volume for all object classes. Each  neural object volume is composed of a mixture of 1281  gaussians. We followed exactly the same approach as the one described in the main paper. We trained our model for 200 epochs using contrastive learning, including the background features.
On top of the results provided in the main paper, we provide some additional 3D pose estimation results in Table \ref{supp:tab:ablation_pose} to compare the performances of both our approach with cuboid and spherical neural object volumes.

\begin{table}[h]
\caption{3D-Pose Estimation results. A prediction is considered correct if the angular error is lower than a given threshold (\ie $\frac{\pi}{6}$, and $\frac{\pi}{18}$). Higher is better. We observe that a spherical neural object volume performs well in IID scenarios but does not generalize as well as the cuboid in more difficult OOD scenarios.}
 \begin{center}
    \tabcolsep=0.11cm
        \begin{tabular}{l|c|cccc|c|cccc}
    \toprule
    Threshold & \multicolumn{5}{c|}{$\pi/6$} & \multicolumn{5}{c}{$\pi/18$} \\
    \hline
    Dataset & P3D+ &\multicolumn{4}{c|}{occluded-P3D+}& P3D+ & \multicolumn{4}{c}{occluded-P3D+} \\
    \hline
    Nuisance& L0 & L1 & L2 & L3 & Mean & L0 & L1 & L2 & L3 & Mean\\
    \midrule
    \OURS-sphere& 90.2 &75.0&62.1&40.1&59.1&65.9&47.0&30.1&13.9& 30.3\\
    \OURS-cuboid& 90.1 &82.6&71.3&52.7&69.1&69.5&55.5&40.6&22.2&39.7\\
    \bottomrule
    \end{tabular}
    \label{supp:tab:ablation_pose}
  \end{center}
\end{table}

\subsubsection{"single feature" ablation}
\label{sec:supp:ablation:single_vertex}
To assess the efficacy of our 3D representation, we conducted an ablation study focused on the 3D representation itself. This involved replacing the neural object volume with a single feature vector. Hence, at the difference of \Cref{sec:method}, we do not define any neural object volume for this case, instead, for each class $y$ we define the feature vector $ C_y' \in \mathbb{R}^D$. Subsequently, we executed a contrastive learning process, randomly selecting positive features from the object in the image. We omitted the background model from consideration since the number of feature vectors aligns with the number of classes (\eg 12 for P3D+). Consequently, these features are already suitably distributed within the feature space and the background clutter model was not necessary. During the classification inference stage (where predicting the 3D pose is not feasible within the current setup), we computed features matching with the 12 feature vectors at our disposal. The predicted class corresponds to the class for which the feature vector exhibits the most significant matching with the image features.

\section{Estimation of concentration parameters}
\label{sup:kappa}
\begin{figure}
    \centering
    \includegraphics[width=\linewidth]{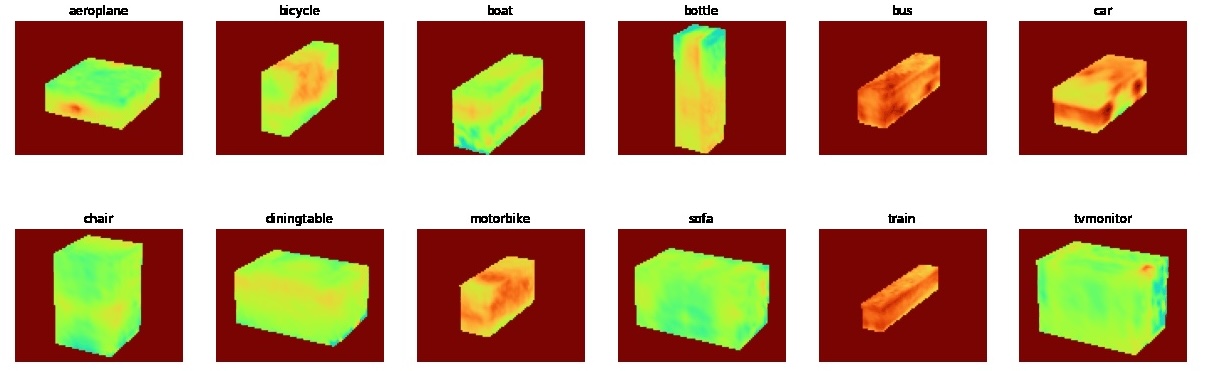}
    \caption{Plot of the concentration $\kappa$ estimates for each vertex using the training dataset (range shown is 0.5 (blue) - 1 (red)).}
    \label{sup:fig:concentration}
\end{figure}

There is no closed form solution to estimating the the concentration ($\kappa$) parameters. Therefore, we set it to $\kappa=1$. To ensure that setting the values to a constant, we tested the effect of approximating the parameter using a standard method as proposed in \cite{sra2012short}. In Figure \ref{sup:fig:concentration}, we can observe that the concentration of the learned features is slightly higher, where the object variability is low (\eg wheels, bottle), whereas it is lower for objects with very variable appearance or shape (\eg airplanes, chairs, sofa). When integrating the learned concentration parameter into the classification and pose estimation inference we observe almost no effect on the results (see Table \ref{sup:tab:kappa}). Importantly, estimating these parameters is non-trivial because of the \textbf{lack of a closed-form solution and potential imbalances among the visibility of different vertices}. This needs to be studied more thoroughly, but our hypothesis is that the learned representation compensates for the mismatch of the cuboid to the object shape and therefore the weighting of the concentration parameter does not have a noticeable effect.

\begin{table}[h]
    \caption{
    Classification accuracy results on PASCAL3D+ and occluded-PASCAL3D+. }
    \label{sup:tab:kappa}
   \begin{center}
  \tabcolsep=0.09cm
    \small
    \begin{tabular}{l|c|cccc}
    \toprule
    Dataset & P3D+ &\multicolumn{4}{c}{occluded-P3D+}\\
    \midrule
    Nuisance & L0 & L1 & L2 & L3 & mean \\
    
    \midrule
    Ours with estimated $\kappa$&99.5&97.2&88.4&59.2&81.6\\
    Ours with $\kappa=1$&99.5&97.2&88.3&59.2&81.6\\

    \bottomrule

    \end{tabular}
    \label{tab:classification_table2}
  \end{center}
\end{table} 

\section{Additional visualizations}
\label{sec:supp:visualizations}

We have generated \textbf{supplementary videos} showcasing the feature matching between image features and the neural object volume features which are available at \url{https://genintel.github.io/NOVUM}.

Additionally, in Figure \ref{fig:supp:qualitative}, we provide qualitative results. Every image is OOD data with different nuisances. We can see that these scenarios are very likely to be encountered by classification models in the real world. Given the class and 3D pose predictions originating from our approach, we overlaid a 3D CAD model on the input image. We can see how our approach successfully predicts both the object category and object pose for these challenging images. 
{\setlength\belowcaptionskip{0pt}
\begin{figure}[h]
    \begin{subfigure} {.32\textwidth}
        \centering
        \includegraphics[width=\linewidth]{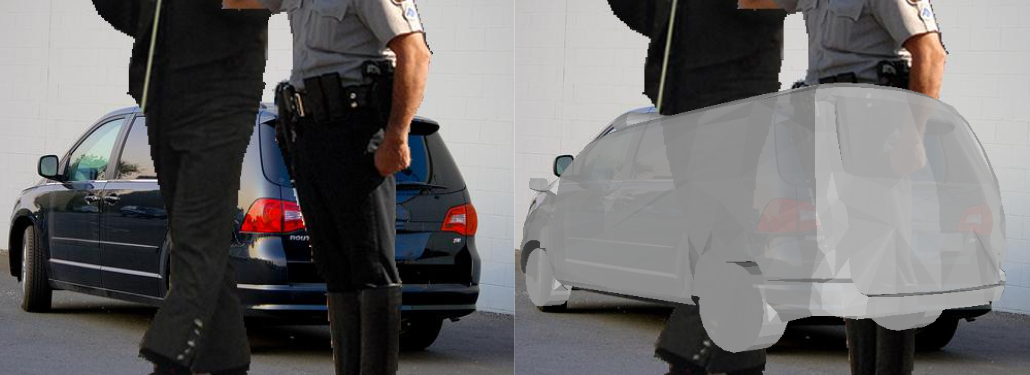}
        \caption{Occlusion (L2), Car}
        \label{fig:L2-figure}
    \end{subfigure}
    \begin{subfigure} {.32\textwidth}
        \centering
        \includegraphics[width=\linewidth]{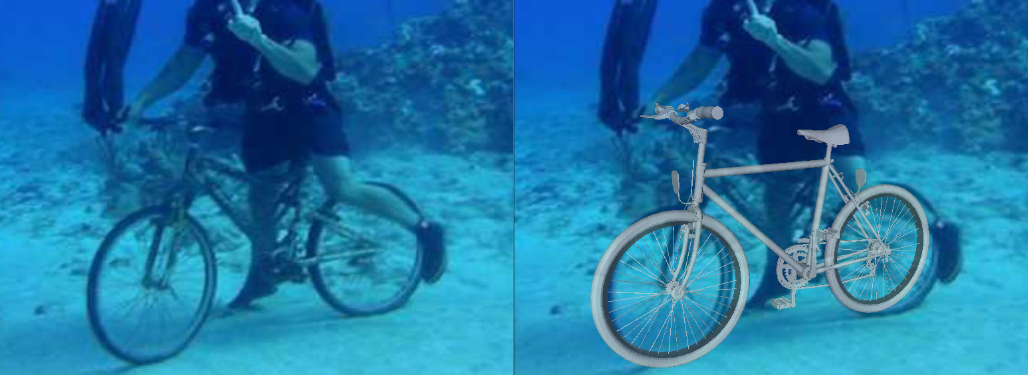}
        \caption{Context, Bicycle}
        \label{fig:context-figure}
    \end{subfigure}
    \begin{subfigure} {.32\textwidth}
        \centering
        \includegraphics[width=\linewidth]{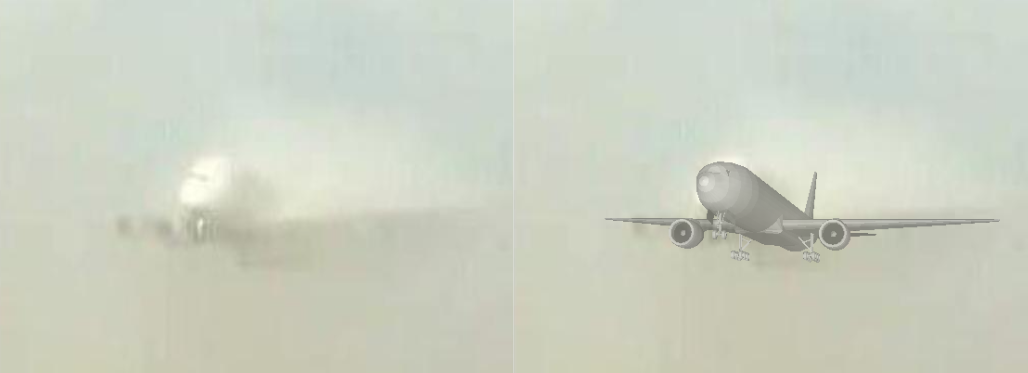}
        \caption{Weather, Plane}
        \label{fig:weather-figure}
    \end{subfigure}
    \newline
    \begin{subfigure} {.32\textwidth}
        \centering
        \includegraphics[width=\linewidth]{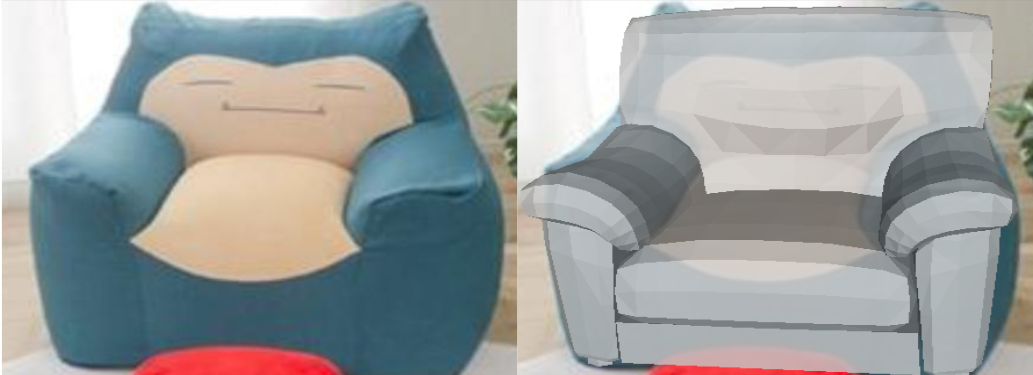}
        \caption{Texture, Sofa}
        \label{fig:texture-figure}
    \end{subfigure}
    \begin{subfigure} {.32\textwidth}
        \centering
        \includegraphics[width=\linewidth]{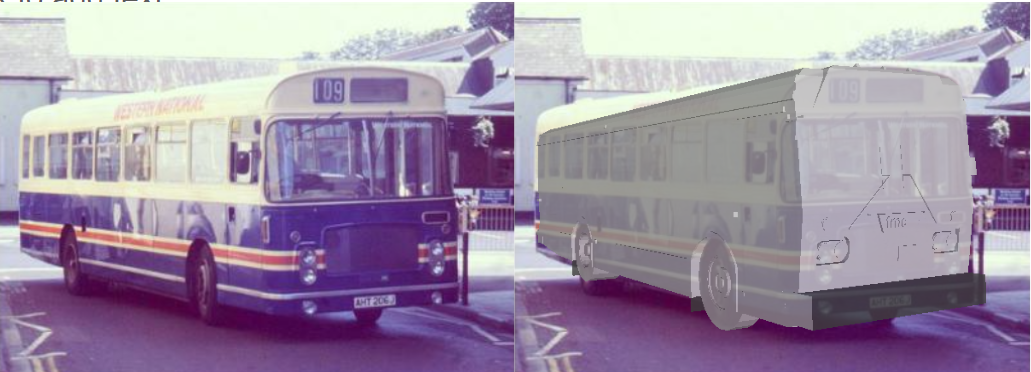}
        \caption{Shape, Bus}
        \label{fig:shape-figure}
    \end{subfigure}
    \begin{subfigure} {.32\textwidth}
        \centering
        \includegraphics[width=\linewidth]{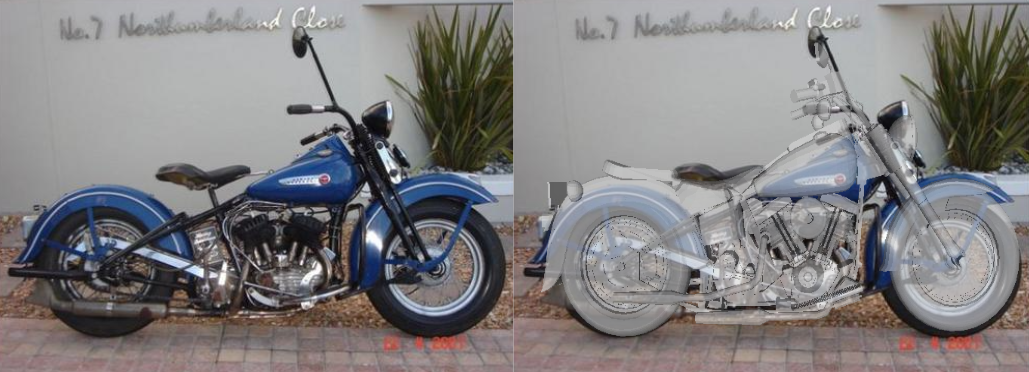}
        \caption{Pose, Motorbike}
        \label{fig:pose-figure}
    \end{subfigure}
    \begin{subfigure} {.32\textwidth}
        \centering
        \includegraphics[width=\linewidth]{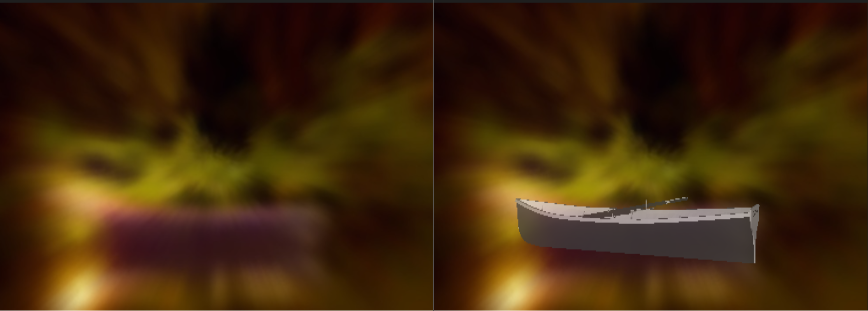}
        \caption{Zoom Blur, Boat}
        \label{fig:zoom-figure}
    \end{subfigure}
    \begin{subfigure} {.32\textwidth}
        \centering
        \includegraphics[width=\linewidth]{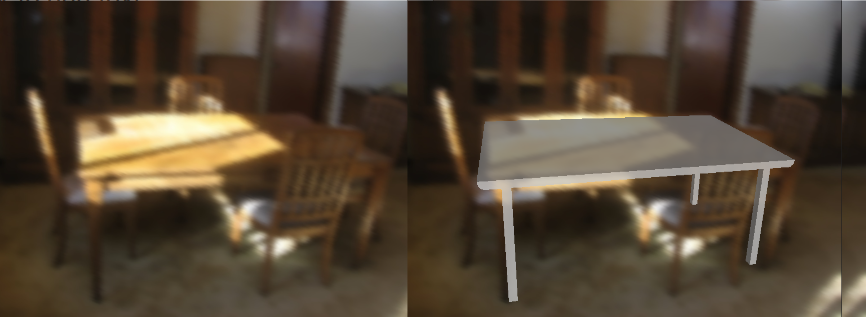}
        \caption{Motion Blur, Table}
        \label{fig:blur-figure}
    \end{subfigure}
    \begin{subfigure} {.32\textwidth}
        \centering
        \includegraphics[width=\linewidth]{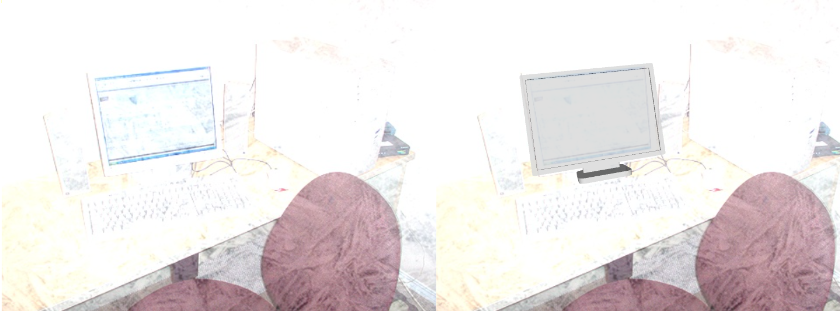}
        \caption{Frost, TV}
        \label{fig:frost-figure}
    \end{subfigure}  
    \begin{subfigure} {.32\textwidth}
        \centering
        \includegraphics[width=\linewidth]{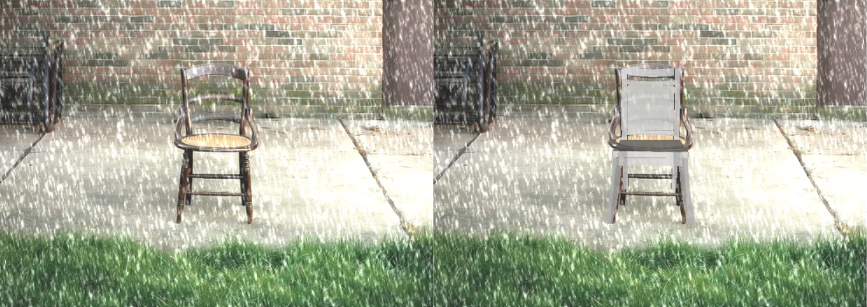}
        \caption{Snow, Chair}
        \label{fig:snow-figure}
    \end{subfigure}
    \begin{subfigure} {.32\textwidth}
        \centering
        \includegraphics[width=\linewidth]{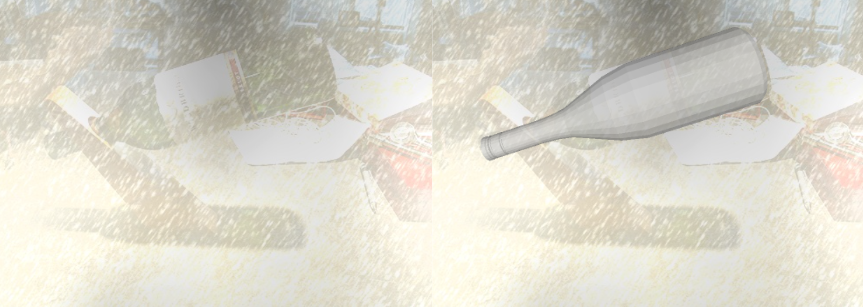}
        \caption{Brightness, Bottle}
        \label{fig:brightness-figure}
    \end{subfigure}
    \begin{subfigure} {.32\textwidth}
        \centering
        \includegraphics[width=\linewidth]{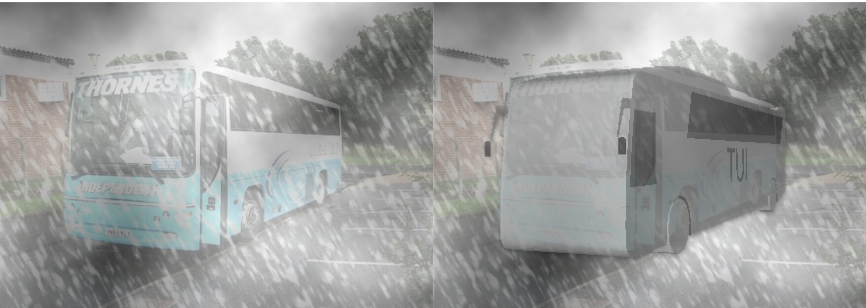}
        \caption{Fog, Bus}
        \label{fig:fog-figure}
    \end{subfigure} 
    \caption{Qualitative results of our approach on Occluded PASCAL3D+ and OOD-CV (a-f), and on Corrupted PASCAL3D+ (g-l). Captions follow the format \textit{\{nuisance\}, \{class\}}. We illustrate the predicted 3D pose using a CAD model. Note that the CAD model is not used in our approach. All images were correctly classified by our approach but incorrectly classified by at least one baseline.}
    \label{fig:supp:qualitative}
\end{figure}
}

\end{document}